# Letting the Data Speak: Extracting Keywords from Crowdsourced Collections with AI

Miguel Arana-Catania*, Catherine Conisbee*°, Matthew Kidd*

## Abstract

Identifying and assigning keywords at scale is a technical, practical, and ethical challenge for crowdsourced collections. This article reports the findings of the "Extracting Keywords from Crowdsourced Collections" project, which used the Their Finest Hour Online Archive, a crowdsourced Second World War digital collection hosted by the University of Oxford, as a case study. The project evaluated three Natural Language Processing approaches to automate keyword extraction: Named Entity Recognition, Keyword Extraction, and Topic Modelling. It tested these approaches across a range of artificial intelligence techniques, from traditional statistical methods to modern GenAI neural networks. Our quantitative and qualitative findings indicate that Natural Language Processing approaches offer real potential for keyword extraction at scale in crowdsourced collections, but that no single method offers a complete solution and that model choice significantly shapes results. We argue that in crowdsourced collections, where metadata is the direct product of engagement with living contributors, automated keyword extraction raises distinct stewardship responsibilities that must be addressed alongside technical performance. Open-weight, extractive models emerge from our evaluation as best placed to support responsible deployment, while generative AI, despite its abstractive potential, introduces accountability risks that anyone managing crowdsourced collections should weigh carefully.

## 1. Introduction

### The problem of keywords

Keywords are central to search and discovery across digital collections.[1] They provide entry points into records, enable filtering and browsing, align with user search behaviour (Ehrmann et al. 2023), and support FAIR principles (Wilkinson et al. 2016) by helping to make data more findable, accessible, interoperable, and reusable. Commonly adopted metadata standards such as DataCite and Dublin Core therefore recommend their

[1] This article uses the term 'keywords' broadly to refer to searchable descriptive terms attached to digital records in order to improve retrieval and discovery. These may include manually assigned tags, automatically extracted terms, named entities, or other searchable subject descriptors. Other literature refer to key terms or keyphrases.

* Equal contribution. University of Oxford, UK
° Corresponding Author: catherine.conisbee@bodleian.ox.ac.uk

inclusion (Salse et al. 2024). Yet identifying, extracting, and assigning keywords at scale presents two interconnected challenges. The first is technical and practical: the process is labour-intensive and time-consuming, often necessitating additional resourcing or automation. This problem is compounded by wider GLAM (Galleries, Libraries, Archives, Museums)-sector challenges: digitised collections have grown faster than institutions' capacity to connect users with specific content (Newman et al. 2010), while insufficient investment, constrained infrastructure, digital skills gaps, and pressure to digitise at scale all influence what metadata can be created and how discovery can be improved (Gosling et al. 2022; Bailey et al. 2024; Gooch et al. 2025; Drabczyk et al. 2025; Korn et al. 2024; Cebr 2025). The second challenge is ethical: keywords necessarily simplify represented content, lack contextual depth and nuance, and, like all collections metadata, are not neutral (Long et al. 2017; Schwartz and Cook, 2002). Those responsible for this process must therefore both make, and be answerable for, decisions about which terms are, and are not, included (Pacheco et al. 2023).

## Our case study

These challenges are particularly pertinent in the context of crowdsourced collections, where metadata is the direct result of engagement with living contributors and imposed terms have the potential to misrepresent or mischaracterise contributors' lived experiences. The Their Finest Hour Online Archive (TFH Archive) is one such collection. Created as a result of a University of Oxford (Oxford) based crowdsourcing project (2022-24) and funded by the National Lottery Heritage Fund, it consists of 2,003 "stories" and 25,000 files – ranging from born-digital memoirs to digitised diaries, letters, photographs, and other items – relating to the Second World War.[2] The TFH Archive was published in June 2024 using the Sustainable Digital Scholarship (SDS) platform (Bodleian Libraries, Oxford), which mandates the use of keywords.[3] To address this requirement and improve searchability within the constraints of tight publication deadlines, a keyword tagging workflow was developed, combining a rules-based Python script, an Imperial War Museum-derived controlled vocabulary, and manual review. While this made more comprehensive keyword assignment possible, it remained a resource-intensive process, and raised concerns that generic external themes, concepts, and identities were being imposed on contributors' stories, rather than letting the data "speak for itself".

## Purpose of the article

This article reports the findings of the "Extracting Keywords from Crowdsourced Collections" (EKCC) project, based at Oxford and funded by Digital Scholarship @ Oxford, which emerged as a direct result of these constraints and concerns.[4] Using the TFH

[2] The project used the term 'stories' because contributions mostly reflect personal memories, anecdotes, and interpretations rather than verified historical records.
[3] https://info.figshare.com/user-guide/figshare-metadata-schema-overview/
[4] https://digitalscholarship.web.ox.ac.uk/

Archive as a case study, EKCC set out to understand the extent to which three Natural Language Processing (NLP) approaches – Named Entity Recognition (NER), Keyword Extraction (KE), and Topic Modelling (TM) – can support the automation of keyword extraction at scale, without compromising responsible data stewardship (Padilla et al. 2023; Belteki et al. 2025; Massey et al. 2023).[5]

The question that EKCC set out to answer encompassed technical, practical, and ethical concerns: if fully manual tagging is not feasible at scale, keywords are required by the publication platform, and imposed terms risk misrepresenting contributors' stories, could artificial intelligence (AI) support automated keyword extraction while allowing terms to emerge from contributors' own words? This article reports and discusses the project's findings, and considers whether and how NLP methods can be applied responsibly in crowdsourced collections contexts, and identifies key considerations for those seeking to implement these approaches in their own collections.

## Findings, arguments, proposals

The EKCC project evaluated NER, KE and TM using multiple methodologies, ranging from statistical methods to neural networks, and in the latter case, from open-source encoders to closed-source generative AI models. These approaches were tested on a sample of TFH archive records. From a **technical** standpoint, we argue that NLP approaches offer real potential for supporting keyword extraction at scale within crowdsourced collections, but no single method offers a complete solution: model choice significantly shapes results. For KE, in our tests, statistical methods obtained high precision but low recall, and neural networks scored lower in both aspects. In the case of NER, neural network models obtained both high precision and recall. EKCC testing indicated that smaller task-specific AI models consistently outperformed larger general ones. One NER model warrants particular mention: GLiNER (Zaratiana et al. 2024), a zero-shot model that allows users to define their own entity categories rather than working within a fixed schema. GLiNER offers a greater degree of flexibility, therefore making it especially well-suited to the varied, unpredictable and context-specific content of crowdsourced collections. Equally, KE and TM approaches are free from fixed-category limitations.

These technical differences have **practical** implications for implementation. Statistical techniques and open-weight models may prove useful for rapid prototyping and experimentation, particularly in contexts with limited computational resources or data privacy requirements, which can be pertinent in some digital collections contexts, as the entire process can be executed offline on a local device. Equally, in terms of the reproducibility and transparency of our analyses, as we can access the inner workings of the implementation, these can be replicated exactly by any other researcher. Closed-source models offered as services by commercial providers are more expensive, but for some particular uses, they can offer better results. However, they do not ensure data privacy, transparency, or reproducibility of outputs, and are vulnerable to technological obsolescence.

[5] https://theirfinesthour.english.ox.ac.uk/home

From a practical perspective, model selection requires assessment of both performance and implementation context. For example, smaller, task-specific open-weights models are a good fit for teams with limited infrastructure or technical resources, and at the same time, in our experiments, they yielded the best results in tasks such as NER. Human review remains necessary across all NLP approaches and model implementations, shifting rather than removing the resource burden as a consequence. Responsible deployment requires enhanced digital literacy, interdisciplinary collaboration, and a willingness to state plainly when a tool is not fit for purpose.

Our findings also point to **ethical** questions. Deploying an AI model in the context of a crowdsourced collection arguably creates a relationship of moral entanglement and vicarious responsibility that cannot be set aside in the pursuit of efficiency. Technology, like metadata, is not neutral (Winner 1986, Chap. 2). Some models cannot be deployed responsibly in certain contexts, and we argue that AI deployers should understand this and again state it plainly rather than qualify it away (Schultz et al, 2025). This is especially important when comparing extractive models with generative AI models. Collections held in trust for living contributors present a distinct stewardship challenge: the people whose experiences are recorded in the archive have a stake in how those experiences are described, surfaced, and made discoverable. Generative AI's abstractive capabilities have the potential to enhance discoverability, producing outputs that go beyond what the text explicitly states. But where stewards of crowdsourced collections are seeking to automate keyword extraction, they should not only assess the performance of models, but also consider whether those deploying them can be meaningfully answerable for the outputs the models produce.

These ethical questions point to a broader conceptual issue about what kinds of discovery different technical approaches make possible. Entity-based discovery privileges formal reference – names, places, dates – over potentially richer interpretive descriptions of experience. The choice between extractive approaches, which draw keywords directly *from* the text, and abstractive ones, which can generate richer contextual descriptions *based on* the text, is therefore a choice about who the collection primarily serves and how. These questions have implications that extend beyond TFH to the wider GLAM sector and anyone responsible for making collections of human experience discoverable.

Underpinning these findings is the claim that crowdsourced collections pose a distinctive and underexamined challenge for keyword extraction. TFH is characterised by the forms of complexity that make such collections both valuable and difficult to process: contributors' own descriptions form the archive's metadata; records range from substantial narratives to brief notes; submissions were made both directly online and through volunteer-run events; and while some contributors were wartime witnesses, others preserved inherited family memories, creating digital memorials with significant sentimental value attached to them (Ramsden 2023; Arthur 2014).

What makes our findings distinctive is that the same team was responsible for all three stages of the process: crowdsourcing contributor testimony, publishing the archive, and now evaluating computational methods for enhancing discovery of it. The technical, practical, and ethical questions that different disciplines may address separately therefore had to be confronted together, by a single team, with real consequences for the living contributors to whom the team have been directly answerable.

All the methodologies implemented in this project have been published in the project's open-source repository and are designed to be reusable on any other text dataset (Arana-Catania et al. 2025). [6]

## Structure

This article is structured in four sections. The first reviews relevant literature and methodological approaches, outlining key debates and identifying gaps and sector needs that this research addresses. The second sets out the methodology of the TFH keyword assignment process and the EKCC project. The third section presents the results of the EKCC project, drawing on both quantitative and qualitative analysis. The fourth section discusses the findings and their wider implications, and offers practical recommendations for professionals working in comparable contexts.

# 2. Literature Review

Across several fields, including information science, computer science, and digital humanities, repeated concerns include how keywords are assigned, by whom, with what consequences, and even what a keyword is – a question that remains surprisingly difficult to answer (Firoozeh et al. 2020).

## What are keywords and why are they useful?

The digitisation of archival, library, and museum collections has dramatically increased the volume of records requiring description, placing the manual tasks of indexing, cataloguing, and metadata creation under considerable strain (Colavizza et al. 2021). As finding aids, catalogues, and documents became digital, that data had to be structured and described in ways that enabled effective search and discovery. In this context, keywords – or key terms, keyphrases, or tags – became important as part of wider indexing systems designed to organise access to information (Shah and Bender 2022).

[6] https://github.com/Digital-Scholarship-Oxford/crowdsourced-data-tools

Rather than entirely replacing earlier analogue cataloguing, subject-heading, and indexing systems used across the GLAM sector, keywords represented a continuation of longstanding practices designed to make stored knowledge searchable and retrievable (Cevolini 2026).

Historically, the selection and assignment of keywords within digital collections has involved labour-intensive manual processes, drawing on controlled vocabularies and classification schemes designed to standardise description and improve retrieval. Subject headings and other descriptive keywords have typically been selected and assigned using human judgement, by archivists, librarians, cataloguers, indexers, even volunteers, requiring decisions about which subjects, themes, events, people, places, or concepts were most significant for future discovery and access.[7] This process rests on an assumption that institutions or individuals should decide in advance which keywords or subject headings matter, and then impose them on historical material in their care.

As digital collections expanded, those responsible for extracting and assigning keywords also had to consider how users actually search for information online. Since the late 1990s, the dominance of keyword-based search engines such as Google has accustomed users to retrieving information through short, discrete terms, typically nouns or named entities such as "Paris" or "World War Two" (Ehrmann et al. 2023). By the early 2010s, this shift prompted debates within libraries and archives over whether traditional subject headings and controlled vocabularies remained necessary in increasingly keyword-driven digital search environments (Gross et al. 2015).

At its most basic, online search depends upon users expressing an information need though keywords or questions, which returns a ranked list of relevant information objects (Shah and Bender 2022). In a GLAM context, a keyword's function can be to "characterize and capture the main topics of a large text data collection or a single document" (Alami Merrouni et al. 2020). At the time of writing, this remains the dominant mode of search across digital collections: As Ehrmann et al. (2023) showed, "80% of search queries on the National Library of France's portal Gallica contain a proper name, and geographical and person names dominate the searches in various digital libraries, be they artworks, domain-specific historical documents, historical newspapers, or broadcasts". The prevalence of this user behaviour at least partly explains why improving keyword coverage remains a priority across the GLAM sector, including at the National Archives (UK) which, in November 2025, announced the introduction of additional subject tags in its online catalogue to "open up the archive to everyone, with any query" (Bell 2025).

Yet expanding keyword coverage across collections manually remains a costly, time-consuming, and difficult-to-scale process. In response, some GLAM institutions have experimented with approaches to supplement traditional manual indexing practices, with broader forms of participation. Crowdsourced, collaborative, or social tagging, and folksonomy-based approaches, invite wider communities of contributors to add keywords and descriptive terms to digital resources, enabling users to discover and navigate materials through language closer to how they themselves search (Kipp and Campbell 2010; Lu et al. 2010). However, social tagging has clear limitations: it depends on sustained user participation, can produce uneven coverage across collections, often requires further standardisation (Poole 2019; Macgregor & McCulloch 2006) and it relies on description flows from individuals and institutions to archival data, rather than from archival data to description (Colla et al. 2022).

[7] Index terms most commonly taken from classification systems such as Library of Congress Subject Headings (LCSH).

## Automation of keywording process

The growing volume of digitised materials has driven the exploration of automated methods, which promise to reduce the resource burden of keyword selection (Colavizza et al. 2021). Automated approaches span several decades, from early statistical methods such as Term Frequency-Inverse Document Frequency (TF-IDF) to machine-learning systems for automated subject cataloguing and metadata generation (Golub et al. 2016; Golub 2021; Jaillant 2022). Over the past 15 years, the British Library has tested a variety of computational, corpus linguistics, and machine-learning methods for working with its large-scale collections (Ridge 2025). Over the same period, the German National Library has used automated keyword assignment as part of wider machine-supported subject indexing, developing platforms based on Annif – an automated subject indexing and classification tool developed by the National Library of Finland (Schumann and Mergel 2026) – to generate Dewey Decimal Classification categories and German Integrated Authority File subject headings/keywords. This extended subject access to materials that would otherwise “not be indexed at all or only partially”, while still requiring librarians to monitor quality and make final judgements (Poley et al. 2025; Mergel et al. 2026).

More recently, studies have explored contextual language models based on transformer architectures, particularly BERT (Bidirectional Encoder Representations from Transformers). BERT-variant architectures have dominated the implementations in NLP tasks overall until the recent rise of GPT-like generative models. Chou and Chu (2022), for example, examined whether BERT-based approaches could support machine-assisted subject indexing in the Project Gutenberg collection, by recommending Library of Congress Subject Headings for digitised texts.

## Named Entity Recognition

Within archival and cultural heritage contexts, much of this work and related scholarship has focused on NLP, a subfield of AI, to support the processing, organisation, analysis, and discovery of digitised collections. NER – the automated identification and classification of named entities such as people, places, organisations, and events within a text – has attracted particular attention (Aejas et al. 2021; Ehrmann et al. 2023). Like TFH, Van Hooland et al. (2015) explored whether NER could be used to identify meaningful concepts within unstructured metadata fields such as “description”, examining the opportunities and limitations of NER to improve discovery. At The National Archives (UK), the “Traces through Time” project showed how identifying and connecting named entities across different archival fonds could support more flexible forms of search and discovery than those offered by catalogue structures alone (Ranade 2016, cited in Colavizza et al. 2021). More recently, Carter et al. (2022) described an interdisciplinary project using AI and machine-learning techniques to extend subject indexing and identify key entities in the Morgenthau Diaries, enabling users to discover “deeper” relations and themes while retaining archival context.

Recent work has continued to push NER beyond standard categories such as people, places, and organisations, showing how domain-specific typologies, manually annotated training data, and refined NER models can identify more collection-specific entities, attributes, and relationships (Luthra et al. 2024; Long and Li 2026). The Congruence Engine project (Boon et al. 2025), for example, fine-tuned a GLiNER model to identify domain-specific entity types in historic textile industry sources. However, bespoke NER tools that rely on custom typologies and manually annotated examples create practical and methodological constraints for crowdsourced collections. Building typologies, annotating training data, and training and validating models requires substantial time, resources, and expertise, while models developed for one archive rarely map cleanly onto another without further adaptation (Keraghel et al. 2024; Ehrmann et al. 2023; Haffenden et al. 2023).

## Keyword Extraction

Alongside NER, keyword or keyphrase extraction (KE) has also been explored as a means of supporting automated description at scale. KE approaches automatically identify and rank the words or phrases considered to be “meaningful”, “representative”, “relevant”, “descriptive”, and/or “expressive” for a document (Merrouni et al. 2020; Giarelis and Karacapilidis 2024). With the development of word embeddings and transformer-based language models, KE approaches can now capture semantic relationships and contextual meaning more effectively, enabling systems to identify keywords and phrases that better reflect document themes and content (Giarelis and Karacapilidis 2024).

## Topic Modelling

TM offers another approach to automated description. Rather than extracting entities or keywords from individual records, it analyses a collection as a whole to identify recurring or “hidden” semantic themes (Abdelrazek et al. 2023), producing sets of representative terms (“topics”) that can guide description and discovery. In GLAM contexts, TM has been explored as a way of generating candidate topics that could function as keywords or subject headings, with human review used to assess and refine the results (Glowacka-Musial 2022).

TM is not new: in 2007, librarians at the University of Michigan tested it on a large corpus of metadata containing 2.5 million records, generating 500 clusters, which were then reviewed by community labellers, with selected labels mapped onto a locally created classification system, based on the Library of Congress Classification System and added to metadata records as subject fields (Hagedorn et al. 2007). More recently, Andresel et al (2023) explored TM for the automated curation of historical documents in Europeana and Transcribathon, finding that topic detection could support the clustering and recommendation of historically related documents, while also identifying challenges around data quality, multilingualism, and topic coherence.

These examples show both the potential and the limits of TM for automated description. TMs can reduce the burden of identifying broad patterns across large collections, but they also tend to privilege the most general and prominent themes in a corpus, potentially obscuring more specific or nuanced topics users may wish to find (Harandizadeh et al. 2022). They also still depend on human interpretation: someone must decide what each cluster represents, which clusters are “meaningful and usable”, and how labels should be mapped onto existing categories (Newman et al. 2010; Churchill and Singh 2022). This introduces familiar problems of human bias, and presupposes that those making these decisions need to have a sufficient knowledge of the collection (Cain 2016) as well as some indication of what users of the collection are interested in finding. Even recent attempts to reintegrate “experts” into the TM process, by allowing their knowledge to guide models (Harandizadeh et al. 2022), or keyword-centric approaches (Pirniakan et al. 2026; Eshima et al. 2024), which rely on “user knowledge” (Wang et al. 2025) and human-generated keywords, merely shift human intervention rather than reducing it.

## Generative AI

Generative AI is also increasingly being explored in automated archival description and discovery. Ridge (2025) notes that LLMs have the potential to “generate structured data by identifying dates, locations, people (and their relationships) and events within narrative texts”, putting NLP techniques such as NER “within reach of researchers without advanced technical skills”. Working with “noisy” user-generated datasets similar in many ways to crowdsourced collections, Hernandez-Camero et al. (2025) found that using LLM-generated explanations alongside TMs could improve topic coherence and, in some cases, produce clusters that aligned more closely with pre-existing labelled categories. And at the Royal Navy Museums (Reusens et al. 2025) found that Llama – a family of large language models developed by Meta – showed promise for keyword generation in a museum collections context, though the study also reported significant limitations including high error rates, hallucination, and bias, and recommends human validation throughout.

## Crowdsourcing and crowdsourced archives

The decisions involved in automated keyword extraction, including the question of what even counts as a useful keyword, become more acute in crowdsourced collections. As Padmavilochanan et al (2025) argued, “extracting structured information from narrative data remains a methodological challenge due to the inherently complex, multilayered nature of human expression.” In crowdsourced collections, which can also be described as “Community Collections” (Berglund Prytz, 2021) or “Community-Generated Digital Content”, (Hughes et al, 2025) there is also an “ethical requirement to honour the work of participants” (Ridge et al. 2021).

With projects such as TFH, crowdsourcing has the potential to not only complete and complement “official” historical accounts, but also to contradict them; to represent and amplify marginalised voices otherwise hidden or absent from existing archival accounts; and to provide the public with access to less familiar stories and alternative perspectives and experiences (Hughes et al. 2025). It is perhaps no surprise then that the value of such collections is well recognised, and in the UK there is a well-established tradition of engaging with communities to record and preserve accounts of events, regional and folk histories, traditions, language, and cultural memory, including of the Second World War (e.g. Mass Observation, The Canna Sound Archive, BBC Domesday Project (1984-1986), BBC People’s Ware Archive). [8] [9] [10] [11]

The TFH project team recognised the significance of their role in safeguarding the stories and objects contributors entrusted to them (Conisbee, 2025), acknowledging the sentimental value attached to many stories and objects shared by living contributors. The need for contributors to retain as much agency as possible – defined by Murray (2017) in her writing on interactivity in digital environments, as “the satisfying power to take meaningful action and see the results of our decisions and choices” – was also at the core of the project’s ethos and decision making, which included “[imposing] minimal restrictions on submissions beyond copyright compliance and relevance to the second world war’ and not censoring or reviewing submissions for historical inaccuracies (Conisbee 2025). Automated keyword extraction in crowdsourced collections therefore raises questions that extend beyond technical opportunities and limitations, and encompass its impact on the subjects and contributors whose stories, memories, and materials are being preserved and made discoverable.

---

[8] https://massobs.org.uk/about-mass-observation/
[9] https://www.nts.org.uk/stories/unlocking-the-secrets-of-the-canna-sound-archive
[10] https://www.computinghistory.org.uk/domesday/
[11] https://www.bbc.co.uk/history/ww2peopleswar/

# 3. Methodology

## 3.1 Problem definition

The EKCC project's methodological choices sought to directly address the limitations of the keyword tagging workflow developed during the TFH project and the concerns it surfaced. As outlined within the Introduction, the TFH archive consists of 2,003 records and was published on the SDS platform, an institutional instance of Figshare, which utilises a metadata schema based on DataCite. While DataCite recommends the inclusion of subject keywords, the SDS platform *mandates it*: each record must contain at least one keyword before publication.[12] Meeting this requirement at scale, for a collection as complex as the TFH archive, presented real challenges.

Contributions to the TFH archive were submitted either directly online (41%) or through 73 volunteer-supported in-person collection events held across the United Kingdom (59%) (TFH. Explore by Collection Day. (n.d.)). Contributors were invited to share their stories in their own words, resulting in metadata field content highly inconsistent in terms of length, detail, and focus. Entries ranged from 10,000 characters (the description field character limit) to little more than a brief note. Submissions entered the archive through different pathways: some contributors catalogued their own material, while others shared their stories with volunteers at in-person events, meaning some records reflect unmediated contributions while others passed through a layer of summarisation. In some cases, the metadata was supplied by volunteers through transcribed interviews, but the file included self-typed first-hand autobiographical accounts. Contributors also differed in their relationship to the events described: some had lived through the war, while others were recounting other people's memories.

To address the SDS platform's keyword requirement, the TFH project developed a keyword tagging workflow which combined a rule-based Python script, controlled vocabulary, and manual review. Where a word in the controlled vocabulary (e.g. "France") appeared in a record's "Description field" or "Item list", the Python script assigned the matching keyword and associated terms defined within the vocabulary (e.g. "French"). This sought to address the inconsistent use of terms like "POW" and "Prisoner of War", and meant that keywords could be applied even when not present in the record metadata. However, this approach conforms to the notion that individuals – in this case, project team members – should decide in advance which keywords matter, and then impose them on historical material. The limitations encountered in this initial approach, using rule-based scripts and a controlled vocabulary, led us to explore the very different NER, TM and KE approaches presented in this article.

[12] https://info.figshare.com/user-guide/how-to-fill-in-the-metadata-fields-first-step/ ; https://info.figshare.com/user-guide/how-discoverable-is-my-research/ ; https://datacite-metadata-schema.readthedocs.io/en/4.6/properties/subject/ ; https://info.figshare.com/user-guide/figshare-metadata-schema-overview/

## 3.2 Dataset

The TFH archive (Lee et al. 2024) is openly available through the SDS platform under a CC-BY-4.0 licence, with the full dataset also accessible via the Figshare API. Records are hosted at the object level, with each published record minted with a versioned DOI. At the start of the EKCC project, the team used the Figshare API to create a static CSV copy of the archive data for analysis. Due to the significant pre-processing that would have been required to standardise file formats, only data contained in the metadata field "Description" was used for testing the models and approaches discussed in this article.

The "Description" field used in this project contained 2,003 entries, with substantial variation in length. Entries had a mean of 243 words, with a standard deviation of 297 words. The 25th, 50th, and 75th percentiles had, respectively, 54, 137, and 307 words, and the longest entry contained 1,786 words. In terms of characters, entries had a mean length of 1,361, and the standard deviation was 1,648. The 25th, 50th, and 75th percentiles were 315, 765, and 1,729 characters, and the longest entry contains 9,863 characters. Across the dataset, the vocabulary contained 25,722 words. The ten most common words, excluding stopwords, were "war", "father", "one", "would", "mother", "went", "contributor", "time", "home", and "family"; the most common bi-grams were "air raid", "years old", "north africa", "end war", "home guard", "contributor father", "royal navy", "air force", "one day", "prisoner war"; and the most common tri-grams were "second world war", "world war ii", "air raid shelter", "first world war", "digital collection day", "royal air force", "shared interview digital", "interview digital collection", "transcription interview attached", "interview attached record".

**Repository locations:**

Their Finest Hour Online Archive (Lee et al. 2024):
URL: https://portal.sds.ox.ac.uk/Their_Finest_Hour
DOI: https://doi.org/10.5287/ora-nzpbbm0v5

Method implementations and accompanying README files (Arana-Catania et al. 2025):
URL: https://github.com/Digital-Scholarship-Oxford/crowdsourced-data-tools
DOI: https://doi.org/10.5287/ora-yqqxywn02

## 3.3 Methods

The EKCC project evaluated three AI approaches to automated keyword extraction: NER, KE, and TM. NER selects the words from each record that belong to a series of predefined categories. KE analyses each record description in the archive independently, selecting the most significant words. TM analyses the entire collection of records simultaneously, sorting them into the most representative groups ("topics"), and then extracting the

keywords from each group. These approaches were selected because each offers different levels of information processing (e.g. at the level of individual sentences or by analysing all records comparatively), imposes different restrictions (e.g. requiring the selection of predefined categories of keywords, or letting the records define their own categories), and offers different perspectives and ways of approaching keyword extraction for crowdsourced collections. Evaluation was carried out comprehensively for each approach, comparing multiple implementations of the same approach to ensure the robustness of the results and to propose different solutions depending on the use case to which it could be applied.

Each of the selected approaches offers a different perspective when it comes to organising the digital archive and defining key terms for its categorisation. KE and TM “let the data speak for itself”, extracting keywords by considering only their use and context. NER starts from specific categories and identifies the words that fit into these categories. Considering them from another perspective, TM begins by analysing the entire collection of records and then connecting the result to each record, while the other approaches analyse the records only individually. From a third point of view, KE and NER select literal words from the archive, while TM extracts strings or words from the archive as an initial guide which generally need to be interpreted and manually assigned a representative textual label. In this way, each methodology prioritises different issues, so it may be more or less applicable to each use case and data set. During the development of the project, we also experimented with combining approaches sequentially.

In terms of implementations, we considered a wide range going from more traditional methods, applying statistical techniques, to more recent methods using neural networks. In the latter set, we applied both open-weight models that can be run on local desktop computers and closed-source models whose execution is offered as a service by private providers (i.e. OpenAI). This not only ensures consistent results but also allows technical aspects to be taken into account when choosing the appropriate methodology for each use case. In the findings section of the Introduction, we described which methods are appropriate in each case, depending on the requirements for replicability, transparency, data privacy and sustainability, or on the basis of the computational resources available. In addition to considering these different aspects, the diverse selection of implementations reinforces the robustness of the conclusions obtained.

Next, we will detail independently each of the approaches used, as well as the implementations chosen, and a general introduction to the categories of neural networks applied.

### 3.3.1 Neural networks

Most of the methods used in NER, KE, and TM involve the use of neural networks trained as large language models. This means that these neural networks have been trained to generate language using large text corpora. The neural networks used employ the transformer architecture. Both this architecture and its use in large language models is currently the most common methodology for text processing.

Depending on how these neural networks are specifically trained, and the specific configuration of their architectures, we can describe three main subcategories of use.

The first subcategory refers to their use in generating numerical representations of texts. The neural network allows a text to be represented as a list of numbers (generally called embedding or vector representation) that encodes its textual properties and allows operations to be performed on it, such as comparing two texts to obtain a numerical value representing their similarity. These representations are generally applied to independent sentences, but can be applied to any set of words. In addition to evaluating the similarity between texts, these numerical representations encode textual information in a sufficiently rich and flexible manner so that they can be applied to multiple language processing tasks.

The second subcategory concerns their use for specific language processing tasks. For example, we can train a neural network to classify texts according to specific categories. In this case, the categories are predefined before the neural network is trained, and only data representative of this categorisation task is used for the training. Here, the network becomes specialised; given a text, its only possible output would be the degree to which the text corresponds to the available categories. Examples of this use can be seen in the NER methodology. These specialised models can be sufficiently efficient while requiring few computational resources, so that they can be run on ordinary computers.

The third subcategory involves their use as generative AI models. In this case, given a text, the network will produce a new text as output. In general, these models are trained to include in their input the definition of the task to be performed, so that they can carry out a multitude of tasks, unlike the previous subcategory, where models are specialised in a single task. These general generative AI models, despite their universality of use, can be as efficient or even more than the previous specialised ones, but they require more computing resources, both in their training and in their execution. Therefore, they are usually executed as services offered by private providers.

### 3.3.2 Keyword Extraction methods

The objective of KE methods is to select keywords from a text. To determine whether a word is considered a keyword, it is compared with the rest of the words in the text. In the methods applied, the comparison is made either by considering the frequency of use of the words or by considering their similarity. In the latter case, to define similarity, the frequency of words in a general text corpus is analysed to identify similarity of use in a generic manner.

### 3.3.2.1 KE – Statistical methods

The first subcategory within these methods comprises the following general methods: TF-IDF, KPMiner (El-Beltagy and Rafea 2010), and YAKE (Campos et al. 2020).

- TF-IDF. This method identifies the importance of words based on their frequency in the document in question (Term Frequency, TF), balancing this measure with their frequency across all documents (Inverse Document Frequency, IDF) to discard common words that are not particularly relevant to the document being analysed.
- KPMiner. This method extends the previous TF-IDF formula by introducing factors that weigh compound terms more accurately and consider the position of words.
- YAKE. The core of this method involves calculating various statistical features that are combined to produce the score of each word as a keyword. These features consider issues such as frequency, casing, position, and context relation.

The second subcategory into which we can organise the applied methods groups together the following graph-based methods: TextRank (Mihalcea and Tarau 2004), SingleRank (Wan and Xiao 2008), TopicRank (Bougouin et al. 2013), PositionRank (Florescu and Caragea 2017), MultipartiteRank (Boudin 2018), RaKUn (Škrlj et al. 2022).

These methods represent the analysed text as a graph in which words are represented as nodes, and the edges are defined by the similarity between them, which at the statistical level is linked to their co-occurrence.

- TextRank. The edges of the graph are initially defined based on the co-occurrence distance between words. The nodes are limited to nouns and adjectives. The TextRank model is applied to this initial graph, by essentially considering each edge as a "vote" between nodes, and repeating these votes recursively until it converges. It is an adaptation of Google's original web search method, PageRank, considering words instead of websites.
- SingleRank. This is an extension of the TextRank method that considers an additional weight on the edges defined as the number of times the connected words co-occur in all documents. This introduces a global correction similar to the IDF factor in the TF-IDF statistical method.
- TopicRank. Unlike the previous methods, in this case the nodes of the graph are topics rather than words. A topic is defined as a cluster of similar words. To obtain the keywords, one word is selected from each topic. In this way, the graph is expected to better represent the semantic relationship between topics.

- PositionRank. A variation of TextRank that weights words by summing the inverse of the positions in which they appear. This gives more importance to words that appear earlier in the document.
- MultipartiteRank. This variation of TopicRank considers multipartite graphs. These allow considering the concept of topics, even though words are used as the main unit. Words are organised into topics, edges are created between words and not topics, but edges between words from the same topic are not permitted.
- RaKUn. The first version of this method (Škrlj et al. 2019) creates meta-vertices by identifying groups of vertices and nodes that can merge together. The second version, used here, merges tokens before constructing the initial graph (Škrlj et al., 2022).

### 3.3.2.2 KE – Neural network methods

In the case of KE, we focused on the first and third subcategories presented in the previous neural network subsection. In the first subcategory, generating numerical representations of the words used in the records to calculate their similarity, we used the following open-weights technique:

- KeyBERT. The specific models used within this technique are the three top SentenceTransformers[13] models trained in sentence similarity. The details of the AI models are as follows: “sentence-transformers/all-MiniLM-L6-v2” (BERT-type teacher-student trained model based on MiniLM model), “sentence-transformers/all-mpnet-base-v2” (BERT-type MPNet model), and “sentence-transformers/paraphrase-mpnet-base-v2” (BERT-type teacher-student trained model based on MPNet). For more details about the models, all the information is on the model card, which can be found on the Hugging Face[14] link of each model.

Within the third subcategory, using generative AI models, we utilised the following closed-weight models:

- OpenAI. The model used was GPT-4o. OpenAI does not publish information about their models, therefore the technical details are unknown.

### 3.3.3 Named Entity Recognition methods

NER methods identify words in a text that belong to a series of predefined categories (e.g. “people”, “location”, “law”, “date”). These methods are therefore not designed to obtain keywords. However, depending on the use case, one may consider, for example, using the top-mentioned entities of certain categories – for example, people or locations – as keywords for the archive.

[13] https://sbert.net/
[14] https://huggingface.co/models

For the EKCC project, we focused on neural network methods, which are the most commonly applied in NER. These include some methods based on more traditional architectures (i.e. Long Short-Term Memory networks) and a wide selection of methods using transformers, as in the previous case of KE. Considering the type of outputs produced by the methods, we can group them into two main categories, which we present below.

### 3.3.3.1 NER – Neural network closed-entity methods

This category includes specialised machine learning methods trained to perform specific language processing tasks, specifically the identification of named entities from a set of predefined categories. The training of these models is performed using datasets such as OntoNotes or CoNLL-2003, which contain texts labelled with the corresponding named entities. Because these datasets consider a set of specific categories, these models can only identify those categories.

**Table 1**. NER models organised according to the entities they can recognise.

| Entity categories | Models |
|---|---|
| **Category A.** "CARDINAL", "DATE", "EVENT", "FAC", "GPE", "LANGUAGE", "LAW", "LOC", "MONEY", "NORP", "ORDINAL", "ORG", "PERCENT", "PERSON", "PRODUCT", "QUANTITY", "TIME", "WORK_OF_ART". | "ner-ontonotes-fast" (Model based on flair embeddings and LSTM-CRF), "ner-ontonotes-large" (Model based on document-level XLM-RoBERTa embeddings and FLERT), "en_core_web_sm", "entities_spacy_en_core_web_lg", "entities_spacy_en_core_web_trf" |
| **Category B.** "ANIMAL", "ARTIFACT", "ASSET", "BIOLOGY", "CELESTIAL", "CULTURE", "DATETIME", "DISCIPLINE", "DISEASE", "EVENT", "FEELING", "FOOD", "GROUP", "LANGUAGE", "LAW", "LOC", "MEASURE", "MEDIA", "MONEY", "ORG", "PART", "PER", "PLANT", "PROPERTY", "PSYCH", "RELATION", "STRUCT", "SUBSTANCE", "SUPER". | "Babelscape/cner-base" (DeBERTa Model by company Babelscape) |
| **Category C**. "corporation", "creative-work", "group", "location", "person", "product". | "tner/roberta-large-wnut2017" (RoBERTa Model by T-NER library), "tner/deberta-large-wnut2017" (DeBERTa Model by T-NER library) |
| **Category D**. "LOC", "MISC", "ORG", "PER". | "ner-large" (Model based on document-level XLM-RoBERTa embeddings and FLERT), "ner-fast" (Model based on flair embeddings and LSTM-CRF). "xlm-roberta-large-finetuned-conll03-english" (RoBERTa Model by FacebookAI, "dbmdz/bert-large-cased- |
| | finetuned-conll03-english" (BERT Model by the Bayerische Staatsbibliothek), "Babelscape/wikineural-multilingual-ner" (mBERT Model by company Babelscape), "elastic/distilbert-base-uncased-finetuned-conll03-english" (DistilBERT Model by company Elastic), "elastic/distilbert-base-cased-finetuned-conll03-english" (DistilBERT Model by company Elastic). |

All methods evaluated in this category are open-weights methods that can be run on standard local computers.

#### 3.3.3.2 NER – Neural network open-entity methods

This category includes methods that allow entity types to be defined each time the method is executed. In this case, we evaluated one open-weights method and one closed-weights method.

To facilitate comparative evaluation with the previous methods, we selected a set of categories equivalent to that used in the previous case: “cardinal”, “date”, “event”, “facility”, “geopolitical”, “language”, “law”, “location”, “monetary”, “affiliation”, “ordinal”, “organization”, “percentage”, “people”, “product”, “measurement”, “artwork”.

- GliNER. The models used were: "EmergentMethods/gliner_medium_news-v2.1", "urchade/gliner_large-v2.1", "knowledgator/modern-gliner-bi-large-v1.0", "gliner-community/gliner_large-v2.5", and "gliner-community/gliner_xxl-v2.5"
- OpenAI. The models used were GPT-4o and GPT-4o-mini. OpenAI does not publish information about their models, therefore the technical details are unknown.

### 3.3.4 Topic Modelling methods

TM methodologies perform an analysis that, unlike the previous ones, is fundamentally global, considering the entire collection of records as a whole, to define a series of topics that represent the main topics covered in them. Once the topics have been defined, each record is classified according to its degree of belonging to each topic. As an additional product of the methodology, each topic is defined by its most representative words. Although this approach does not seek to produce keywords, the most representative words of each topic can be considered good examples of keywords. However, in practice, these terms are not always satisfactory, so it is common for them to be used as an initial guide to define more precise and clear terms.

### 3.3.4.1 TM – Statistical and matrix methods

In this set of methods, we evaluated two main subcategories: statistical and matrix factorisation methods. The first of these includes statistical methods that evaluate the frequency of use of each word in the archive and, based on this data, adjust a probability function with independent parameters for each topic that represents the observed word frequencies. The method used was as follows:

- LDA (Blei et al. 2003; Hoffman et al. 2010). The model considers that the set of records can be generated probabilistically from an initial (multinomial) distribution that models the topics, and once a topic has been sampled, another distribution (in this case Dirichlet) that models the terms corresponding to this topic. Following this model, we can use the record in reverse to adjust the probability distributions, thereby obtaining information on the relationship between words and topics, and between topics and records.

The second subcategory includes matrix factorisation methods. In this case, the archive is represented as a matrix in which records are connected to words, and factorisation techniques are applied to represent this information as the product of two independent matrices, one connecting records to topics and the other connecting topics to words. The first allows us to organise the records into topics, while the second allows us to identify the words representative of each topic. The methods used were:

- NMF (Hoyer 2004; Lee and Seung 2000; Paatero and Tapper 1994) . Non-negative matrix factorisation. A relevant feature of this factorisation algorithm is that it ensures that the matrices representing the conversion of words to topics, or topics to records, do not contain negative terms. This is essential in order to interpret these concepts (e.g. interpreting a record as the combination of certain positive percentages corresponding to each topic).
- SVD. Singular value decomposition provides an alternative algorithm for matrix factorisation.

### 3.3.4.2 TM – Neural network methods

These methods use neural networks to produce representations of each record (in the same way as was done in the KE methods), and compare the joint representation of each record with others, grouping the ones that are more similar. Each of these groups is a topic.

For the EKCC project, we evaluated one open-weights method and one closed-weights method.

- BERTopic. The models used were: "all-MiniLM-L6-v2", "BAAI/bge-m3", "sentence-transformers/paraphrase-multilingual-mpnet-base-v2", "ibm-granite/granite-

embedding-125m-english", and "ibm-granite/granite-embedding-278m-multilingual".

- OpenAI and BERTopic. The OpenAI embeddings model used was text-embedding-3-large, which then was used in the BERTopic. Unlike in the KE and NER approaches, in this case the OpenAI model is not used as a GenAI model, but to generate embeddings. OpenAI does not publish information about their models, therefore the technical details are unknown.

## 3.4 Implementation

In this section, we describe the implementation of each of the methodologies used in the EKCC project: KE, NER, and TM. The implementations described below can be found in the project's open-source repository (Arana-Catania et al. 2025). [15]

### 3.4.1 Keyword Extraction implementation

The keyword extraction methodologies used, as introduced in the Methods section, and their implementations, are summarised in the next table.

**Table 2**. KE implementations

| Method category | Methods |
|---|---|
| General statistical methods | pke library[16] (Boudin 2016): TF-IDF. KPMiner<br><br>YAKE library[17] (Campos et al. 2018a; Campos et al. 2018b; Campos et al. 2020): YAKE |
| Graph-based statistical methods | pke library[18] (Boudin 2016): TextRank, SingleRank, TopicRank, PositionRank, MultipartiteRank<br><br>RaKUn 2.0 library[19]: RaKUn (Škrlj et al. 2022) |
| Open-weight neural network methods using text embeddings | KeyBERT[20] (Grootendorst 2021) using the following Hugging Face models: "all-MiniLM-L6-v2", "all-mpnet-base-v2", and "paraphrase-mpnet-base-v2", implemented using the SentenceTransformers library (Reimers and Gurevych 2019).[21] |
| Closed-weight neural network methods using GenAI. | OpenAI[22]: model GPT-4o |

[15] https://github.com/Digital-Scholarship-Oxford/crowdsourced-data-tools
[16] https://github.com/boudinfl/pke
[17] https://github.com/LIAAD/yake
[18] https://github.com/boudinfl/pke
[19] https://github.com/skblaz/rakun2
[20] https://github.com/MaartenGr/KeyBERT
[21] https://sbert.net/
[22] https://github.com/openai/openai-python

### 3.4.2 Named Entity Recognition implementation

The different approaches to NER that we have used, introduced in the Methods section, and their implementations, are summarised in the following table.

**Table 3**. NER implementations.

| **Method type** | **NER categories** | **Models** |
|---|---|---|
| Neural network closed-entity | NER category A | SpaCy[23] (Honnibal et al. 2020): "en_core_web_sm", "en_core_web_lg", "en_core_web_trf"<br><br>Flair[24] (Akbik et al. 2018): "ner-ontonotes-fast", "ner-ontonotes-large", |
| Neural network closed-entity | NER category B | Hugging Face: "Babelscape/cner-base" |
| Neural network closed-entity | NER category C | Hugging Face: "tner/roberta-large-wnut2017", "tner/deberta-large-wnut2017" |
| Neural network closed-entity | NER category D | Flair (Akbik et al. 2018)[25]: "ner-large", "ner-fast".<br><br>Hugging Face: "xlm-roberta-large-finetuned-conll03-english", "dbmdz/bert-large-cased-finetuned-conll03-english", "Babelscape/wikineural-multilingual-ner", "elastic/distilbert-base-uncased-finetuned-conll03-english", "elastic/distilbert-base-cased-finetuned-conll03-english" |
| Neural network open-entity open-weights | Open-entity | GliNER (Zaratiana et al. 2024).[26] using the following Hugging Face models: "EmergentMethods/gliner_medium_news-v2.1", "urchade/gliner_large-v2.1", "knowledgator/modern-gliner-bi-large-v1.0", "gliner-community/gliner_large-v2.5", "gliner-community/gliner_xxl-v2.5" |
| Neural network open-entity closed-weights using GenAI | Open-entity | OpenAI[27]: models GPT-4o and GPT-4o-mini |

### 3.4.3 Topic Modelling implementation

In the following table, we summarise the TM approaches used and their implementations. The methods are described in the previous Methods section.

[23] https://spacy.io/
[24] https://flairnlp.github.io
[25] https://flairnlp.github.io
[26] https://github.com/urchade/GLiNER
[27] https://github.com/openai/openai-python

**Table 4**. TM implementations

| Method category | Method | Libraries |
|---|---|---|
| Statistical and matrix methods | LDA (Blei et al. 2003; Hoffman et al. 2010), NMF (Hoyer 2004; Lee and Seung 2000; Paatero and Tapper 1994), SVD | scikit-learn library (Pedregosa et al. 2011).[28] |
| Neural network open-weights | "all-MiniLM-L6-v2", "BAAI/bge-m3", "sentence-transformers/paraphrase-multilingual-mpnet-base-v2", "ibm-granite/granite-embedding-125m-english", "ibm-granite/granite-embedding-278m-multilingual". | BERTopic library[29] (Grootendorst 2022) and Hugging Face models |
| Neural network closed-weights | OpenAI embedding model text-embedding-3-large | BERTopic library [30](Grootendorst 2022) and openai library[31] |

Unlike in the KE and NER cases, in TM the neural network closed-weights OpenAI model is not used as a GenAI model, but to generate embeddings. The embeddings were obtained using the openai Python library. Afterwards, those embeddings were used to perform topic modelling using the BERTopic library.

## 3.5 Quantitative evaluation methodology

The two main techniques evaluated quantitatively were KE and NER. Since both techniques select words from each record, it is possible to define metrics that evaluate the degree of matching with the word selections made by humans in a sample of the dataset. In the case of TM, it is more difficult to interpret the definition of what each topic is, as it is defined by the complete set of records for each topic, and thus a dataset sample cannot be evaluated, and by a series of terms that only act as a guide. For this reason, for this latter approach, we have only carried out a qualitative evaluation.

To evaluate the KE and NER models, we created a set of gold standard outputs – manually curated lists of keywords and entities that project team members identified in the sample input text, which served as the benchmark against which the models' performance was measured. The human-selected keywords were all present in the dataset, and thus there were no "absent keywords" in the gold standard. The evaluation was carried out by two evaluators from the team on a sample of ten records. Where disagreement arose, these were resolved through discussion to reach consensus. It is acknowledged that the sample size provides indicative rather than definitive findings, and future research to extend this evaluation to a larger sample is encouraged.

[28] https://scikit-learn.org
[29] https://maartengr.github.io/BERTopic/index.html
[30] https://maartengr.github.io/BERTopic/index.html
[31] https://github.com/openai/openai-python

In terms of our approach to quantitative evaluation, we assessed the models against several performance metrics, which considered different evaluation procedures useful for our project. We therefore recommend a detailed reading of each definition. For example, unless otherwise specified, the entity category “date” has been excluded from the NER evaluation as it was deemed not useful for this project. In some metrics, a direct comparison has been made with the ground truth produced by human evaluators, while in other cases we have carried out a detailed evaluation of the results produced by the models, assessing their correctness. Some metrics only consider a binary evaluation (e.g., in NER, whether the word is identified as an entity or not), while others consider a multi-class evaluation (e.g. in NER, whether the identified entity category is correct: “person,” “location,” etc.). For example, binary evaluation in NER is appropriate if entities are considered as project keywords, as these are not organised by category. With these different considerations, we hope to capture aspects of practical relevance for this project.

To define the metrics used, it is necessary to first define the following quantities:

- True positive (TP). A word selected by the model as a keyword or entity (I.e. a “positive”), which also was selected in the human evaluation (i.e. “true”).
- False positive (FP). A word selected by the model as a keyword or entity (I.e. a “positive”), which however was not selected in the human evaluation (i.e. “false”). i.e. there is disagreement between the AI and the human evaluation.
- True negative (TN). A word that was not selected by the model as a keyword or entity (I.e. a “negative”), which also was not selected in the human evaluation (i.e. “true”).
- False negative (FN). A word that was not selected by the model as a keyword or entity (I.e. a “negative”), which however was selected in the human evaluation (i.e. “false”). i.e. there is disagreement between the AI and the human evaluation.

Below we define details and metrics for each approach.

### 3.5.1 Quantitative evaluation methodology – KE

For each output of each keyword extraction technique, the five most relevant keywords were evaluated. The implementations applied not only select keywords but also allowed them to be sorted by order of importance.

The following metrics were used:

- **Keyword Identification Precision @ 5**: the proportion of the top five words highlighted by the model as keywords that were correct after their human assessment. I.e. TP/(TP+FP).
- **Keyword Identification Recall @ 5**: the proportion of keywords present in the human gold standard, that were correctly highlighted as being in the top five keywords by the model. I.e. TP/(TP+FN).

### 3.5.2 Quantitative evaluation methodology – NER

To evaluate the NER implementations, we assessed the following metrics:

- **Entity Identification Precision**: the proportion of words highlighted by the model as entities that were correct after their human assessment (regardless of the entity class being correct or not). I.e. TP/(TP+FP).
- **Entity Identification Recall**: the proportion of entities present in the human gold standard, that were correctly highlighted as entities by the model (regardless of the entity class being correct or not). I.e. TP/(TP+FN).
- **Entity Categorisation Precision**: the proportion of words highlighted by the model as entities of a specific category (e.g. "person", "location",...) that were correct after their human assessment. I.e. TP/(TP+FP) where in this case True and False also consider the correct category. In this metric, all the categories were considered, including the "date" category.

In all cases, partial returns were treated as incorrect. For example, "Liverpool Street" was incorrect whereas "Liverpool Street station" was correct.

In addition to these metrics, we calculated a statistical parameter that we consider to be of practical use for others undertaking similar analysis:

- **Extraction Ratio**: the proportion of the number of entities extracted by the model compared to the number of entities in the gold standard, independent of whether they were correct.

Unlike the previous ones, this parameter is not a performance metric. Its purpose is to provide an approximate measure of the volume of entities highlighted by the different models. These entities need to undergo human validation before being used on a production platform. Therefore, it is informative to know whether the extent of this validation work is similar for all models.

As presented above, each of the NER models works with different sets of entity categories, due to the dataset used for their training, ranging from 4 to 29 categories. Therefore, the results include the entity category according to these sets. The different categories are not strictly comparable, as they perform different tasks.

## 4. Results

In this section, we report the results of this project. These are organised according to quantitative and qualitative evaluations.

### 4.1 Results – Quantitative

The quantitative results are presented in Tables 5 and 6. As set out above, the qualitative evaluation focuses only on the KE and NER methodologies.

**Table 5**. Quantitative results for KE experiments

| Keyword Model | Keyword Identification Precision @ 5 | Keyword Identification Recall @ 5 |
|---|---|---|
| pke_multipartiterank | 0.96 | 0.56 |
| pke_topicrank | 0.94 | 0.49 |
| pke_positionrank | 0.86 | 0.58 |
| pke_kpminer | 0.79 | 0.08 |
| pke_singlerank | 0.75 | 0.58 |
| openai_gpt-4o | 0.72 | 0.39 |
| pke_tfidf | 0.70 | 0.25 |
| pke_textrank | 0.67 | 0.62 |
| pke_yake | 0.49 | 0.40 |
| yake_yake | 0.42 | 0.33 |
| rakun_rakun | 0.24 | 0.06 |
| keybert_all-mpnet-base-v2 | 0.23 | 0.24 |
| keybert_all-MiniLM-L6-v2 | 0.22 | 0.22 |
| keybert_paraphrase-mpnet-base-v2 | 0.16 | 0.14 |

**Table 6**. Quantitative results for NER experiments

| NER Model | Entity category | Entity Identification Precision | Entity Identification Recall | Entity Categorisation Precision | Extraction Ratio |
|---|---|---|---|---|---|
| flair_ner-large | D | 0.96 | 0.89 | 0.93 | 1.08 |
| spacy_en_core_web_trf | A | 0.96 | 0.86 | 0.92 | 1.13 |
| flair_ner-ontonotes-large | A | 0.96 | 0.85 | 0.88 | 1.16 |
| flair_ner-fast | D | 0.95 | 0.87 | 0.91 | 1.05 |
| gliner_knowledgator/modern-gliner-bi-large-v1.0 | A | 0.94 | 0.76 | 0.93 | 0.79 |
| gliner_urchade/gliner_large-v2.1 | A | 0.91 | 0.68 | 0.88 | 1.10 |
| openai_gpt-4o | A | 0.90 | 0.85 | 0.82 | 1.30 |
| spacy_en_core_web_lg | A | 0.88 | 0.80 | 0.81 | 1.13 |
| hf_xlm-roberta-large-finetuned-conll03-english | D | 0.87 | 0.85 | 0.86 | 1.16 |
| flair_ner-ontonotes-fast | A | 0.87 | 0.83 | 0.85 | 1.11 |
| spacy_en_core_web_sm | A | 0.87 | 0.78 | 0.69 | 1.05 |
| openai_gpt-4o-mini | A | 0.86 | 0.81 | 0.75 | 1.35 |
| hf_Babelscape/wikineural-multilingual-ner | D | 0.84 | 0.82 | 0.79 | 1.11 |
| hf_dbmdz/bert-large-cased-finetuned-conll03-english | D | 0.82 | 0.83 | 0.83 | 1.16 |
| gliner_EmergentMethods/gliner_medium_news-v2.1 | A | 0.82 | 0.61 | 0.90 | 0.81 |
| hf_Babelscape/cner-base | B | 0.72 | 0.87 | 0.62 | 1.96 |
| hf_elastic/distilbert-base-uncased-finetuned-conll03-english | D | 0.61 | 0.61 | 0.59 | 1.36 |
| hf_elastic/distilbert-base-cased-finetuned-conll03-english | D | 0.42 | 0.61 | 0.43 | 1.67 |
| hf_tner/roberta-large-wnut2017 | C | 0.17 | 0.82 | 0.86 | 1.19 |
| hf_tner/deberta-large-wnut2017 | C | 0.17 | 0.78 | 0.84 | 1.03 |

In the KE experiments, we evaluated keyword identification using precision @ 5 and recall @ 5. With regard to keyword identification precision, the best results were obtained using statistical methods. In particular, MultipartiteRank, TopicRank and PositionRank. The first non-statistical method we encounter as we move down the table is the generative AI neural network model GPT-4o, although it ranks below five statistical methods. The other neural network methods explored, which use numerical representations of words to calculate their similarity, all rank at the bottom of the results.

The recall values paint a similar picture. The top six results are statistical methods. Those achieving the highest scores were TextRank, PositionRank, SingleRank, and MultipartiteRank. Once again, the next neural network method is GPT-4o. The other neural network methods also rank low in the table, although in this case above other statistical methods.

Statistical methods therefore achieved the best results in the KE task. These methods are very simple to implement, can be run locally, do not require significant computational resources, can be scaled to large datasets, and are explainable.

For the NER task, as with the KE task, the first two metrics were precision and recall. For this task, all the methods used are neural network methods. Several of these methods achieve nearly maximum values for the metrics, demonstrating that these techniques are well-suited to the NER task. In terms of entity identification precision, the ner-large, ner-ontonotes-large, and ner-fast models from the Flair library, alongside the en_core_web_trf model from the SpaCy library, occupy the top positions. These were followed by two GliNER models, and then by GPT-4o. The entity identification recall results follow a similar pattern, with the exception of the cner_base model, which also appeared in the top group.

For entity categorisation precision, the top positions were again occupied by the ner-large, ner-fast, and ner-ontonotes-large models from the Flair library, as well as the three GliNER models and the en_core_web_trf model from the SpaCy library. In this case, however, GPT-4o no longer appeared among the top-performing models.

Across these three metrics, the open-source model downloaded from Hugging Face, which consistently occupied the top positions in the results, excluding the aforementioned Flair and SpaCy libraries, was xlm-roberta-large-finetuned-conll03-english.

The final parameter in the table is the extraction ratio, which is not a performance metric but a parameter that measures the volume of entities provided by the models. The cner-base model produced the highest number of entities, while GPT-4o, ner-ontonotes-large, and xlm-roberta-large-finetuned-conll03-english also ranked highly for this parameter.

## 4.2 Results – Qualitative

Across all three approaches, we found that NLP methods offer distinct opportunities that manual keywording cannot easily replicate, though each also carries limitations that shape what can be surfaced from the data. Understanding these distinct limitations matters when selecting an approach for a given collection and context. In short, no single NLP approach offers a complete solution; the right choice depends on practical context, technical and compliance requirements, and the characteristics of the collection.

### 4.2.1 Results – Qualitative – NER and KE

Since NER and KE extract words directly from the text, these approaches produced **specific, text-grounded keywords** rather than generic, abstract, or vague descriptors. This is significant because manual keywording often requires a degree of abstraction in order to remain manageable, leading to the use of broad categories that can flatten distinctions within the data. By contrast, NER and KE models return terms that are closer to the language used in the text itself. For example, identifying “Land Girl” where it appears in the text, rather than assigning a broader category such as “Women’s War Work” when it does not appear in text, preserves historical and descriptive precision.

Another important application of NER and KE is the **identification of relationship terms within the data**. Much of the TFH material refers to individuals indirectly, through relationships rather than names. For example, a contributor might refer to “my mother” without naming her. Approaches before the advent of neural networks would struggle to capture this, as there is no explicit entity to tag. Some NER models, however, were able to extract relational terms such as “mother” or “brother”. This creates an opportunity to surface individuals who would otherwise remain hidden in the data, particularly women, who, based on our knowledge of the TFH archive, are more likely to be referenced relationally (e.g. “and his wife”).

The ability to surface terms beyond more conventional named entity categories also points to the value of more **flexible NER models, such as GLiNER**. Unlike many NER models that restrict users to different fixed sets of categories, GLiNER – which has several open-source implementations – lets users define custom categories without requiring fine-tuning. This is particularly valuable in crowdsourced collections, where the data is wide ranging. In practical terms, this means that a project team can decide that “POW camp”, “civilian occupation”, or even “religious affiliation” are relevant categories, and instruct GLiNER to extract any words or phrases relating to them. The opportunity therefore lies in the ability to align the process of extracting keywords with the priorities of a particular project, rather than adapting those priorities to fit the limitations of the tool.

The analysis also identified several limitations of the NER tools tested. A fundamental constraint of NER models, including GLiNER, is the reliance on predefined categories that determine in advance what category of words the model is capable of identifying. If a word or phrase does not fit within the available categories, it will not be extracted, regardless of its meaningfulness or potential usefulness for users. At the same time, it should be noted that this limitation does not apply to the KE approaches, and that in the NER case, the number and nature of predefined categories vary considerably from model to model, which will influence which model is most appropriate for a given use case.

A related issue is the presence of potentially **unhelpful or low-value categories**. Many NER models extract entities such as dates and numbers, which may be technically correct, but might add little interpretive value in certain contexts. In the case of NER models, the problem of **ambiguous classification** is a further limitation in how models assign meaning. For example, when a term could reasonably belong to more than one category, the NER model selects only one: "Dunkirk" might only be classified as a "location", even though in historical terms it can also refer to a specific event. This classification is important as it may determine if and how the term is used in search and analysis.

More broadly, the findings point to **imperfect model performance and the need for human validation**. No model tested achieved perfect precision and recall scores. Errors included assigning the wrong category (e.g. categorising "Bailey Bridges" as a "Person" rather than as "Facility"), partial extraction (capturing only part of a phrase, such as "Palestine" rather than "Palestine police"), and omission (failing to identify relevant terms). These errors mean that outputs cannot be used without manual review, which introduces an additional layer of work in the form of validation, which must be accounted for in practice.

### 4.2.2 Results – Qualitative – TM

TM offers the opportunity of **identifying higher-level concepts across the collection**. Manual keywording typically operates at the level of individual records, making it difficult to identify patterns that span the dataset as a whole. TM addresses this by grouping records based on shared language use, thereby revealing recurring themes. For example, clusters of words relating to evacuation, domestic life, or military service may emerge across multiple records. Using these patterns to support thematic tagging or guide further analysis may be useful in large datasets where such patterns would be difficult to identify manually.

At the same time, TM has several limitations, particularly in relation to the **interpretability of its outputs**. The clusters of words it produces are not self-explanatory, and require interpretation and labelling before they can be used. This can make the outputs difficult to work with, especially for users without technical expertise. And as users must decide how to interpret each cluster, different users may assign different labels to the same set of words, leading to inconsistency. The project team found that generative AI can assist here by suggesting labels based on the words within each cluster, which would reduce the burden of manual interpretation and improve consistency, particularly when working at scale. However, it would need to be balanced against the now-familiar risks associated with relying on AI-generated outputs, including hallucinations, explainability, and reproducibility (Huang et al. 2025; Adel & Alani 2025; Alansari & Luqman 2025).

The **learning curve associated with TM** presents another practical limitation. Understanding how to generate, interpret, label, and then apply TM outputs requires a level of technical knowledge and interpretive judgement that may not be available in all project contexts. This can limit the accessibility of the method, particularly in resource-constrained environments. That said, topic modelling is not without value in this context, particularly in combination with other approaches, such as using NER to remove personal names before modelling, or in more analytical and background applications such as recommender systems, where outputs do not need to be directly legible to end users.

## 5. Discussion

The evaluation of the three approaches presented in this article indicates that no NLP approach offers a perfect solution for extracting keywords from crowdsourced collection data. NER, KE, and TM each have advantages and limitations, with all models tested having produced errors. Appropriate method and model selection, therefore, depends on several factors: performance, the characteristics of the collection, practical and technical and ethical considerations, and institutional constraints.

### No single method is perfect: choices matter

The findings of EKCC exemplify the need to critically assess the technologies used to support keyword extraction, to judge if, when, and how they can be applied effectively and responsibly (Newman-Griffis 2025). This is particularly important in crowdsourced collections, where technological choices directly shape the representation of contributors' lived experiences. As Ridge et al. (2023) warn, inadequate training and guidance risks "people unnecessarily using technologies like machine learning, or adopting a technology or method that is inappropriate for their project" and "unaudited biases in AI / machine learning tools introducing errors or inappropriate data".

The results of the NER model testing illustrate the importance of model choice. Although several models performed strongly, performance varied considerably across those tested. This type of variability matters at scale: choosing a model at random may result in more errors than correct results. Selecting an appropriate model for a given institutional context and collection requires an understanding of both the data – including constraints such as text length, language, copyright, and sensitive content – and the specific capabilities and limitations of different models. Navigating this is not straightforward, and it speaks to the value of digital and AI literacy within institutions, of small-scale testing and pilot projects, and of collaboration with technical experts where possible. It also reinforces the need for "humans-in-the-loop" (Taka et al. 2023; Branley-Bell et al. 2025; Lazaros et al. 2026) to ensure the accuracy and integrity of digital collections data.

Although commercially available models such as the OpenAI GPT family have become increasingly popular, and may therefore be a more obvious choice, EKCC testing showed that GPT-4o and GPT-4o-mini were outperformed in the NER task across all measures by the strongest performing Flair and SpaCy models (although further testing with the most recent GPT models would be valuable). Beyond this, FLAIR and SpaCy models have the distinct advantage of being freely available, open models, which can be deployed entirely "locally", meaning that data never leaves an institution's infrastructure. This is meaningful for collections containing copyrighted, licensed, sensitive, or otherwise restricted material; it also means that these models can be deployed within online coding environments such as Google Colab, if appropriate. From a digital preservation standpoint, free and open models also reduce dependency on commercial models over the short, medium, and long term.

## Extraction, abstraction, and generative AI

Beyond quantitative and qualitative findings, the project raises a series of broader conceptual questions about what keywording means and how discovery should be undertaken in crowdsourced historical collections. One discussion that emerged concerns the difference between extracting what is written and interpreting what it refers to, that is, the distinction between extractive and abstractive keywords (Maragheh et al. 2023). Most NER and KE tools directly extract the words present in text input. In historical material, however, many of those words might point to the same events, experiences, or relationships that are not captured by their literal form. This is particularly the case in crowdsourced collections, where spelling, formatting, and terminology are highly inconsistent. For example, "6 June 1944", "6/6/44" and "June 6th 1944" may all refer to D-Day, just as "dad", "pa" and "daddy" can refer to the same family role. Treating these expressions as entirely separate keywords would mirror the inconsistency of the metadata, even though for many users, terms such as "D-Day" and "Father" would be useful entry points to the relevant records.

This points to a related issue about what kinds of content are visible in crowdsourced collections. In an archive such as TFH, contributions often take the form of personal and family stories, which means that much of what matters to users is expressed through informal and emotional language. When engaging with the archive, the project team encountered a wide variety of themes, ranging from cats to Catholicism. As a Second World War archive, TFH also features numerous volunteered descriptions of fear, camaraderie, grief, jubilation, and boredom in indirect and personal ways, none of which map cleanly onto entity categories. These themes are not entities in the standard sense, and they are rarely named explicitly in the metadata, meaning that they were not picked up or used by the tools and models that we tested. Consequently, discovery of archival content built around entities tends to privilege formal references ("the Blitz") over lived experiences ("ran towards their Anderson shelter"), even though users – especially researchers working in the humanities – may find it useful to be able to group, compare, and discover material based on such themes.

One option could be for the original entities extracted from the data (extractive keywords) to sit alongside higher-level umbrella terms or concepts (abstractive keywords). For example, "farmer", "land girl", and "Women's Timber Corps" (which appear in the text) could all remain visible, while "agriculture" (which does not appear in the text) would be assigned to records in which these variants appear. But while this may enhance the discoverability of records that contain content relating to the broader, abstracted theme, this approach would raise several issues for crowdsourced collections such as TFH. Creating and maintaining the lookups and mappings needed to support this kind of layering would require time, expertise, and sustained resources that most crowdsourced projects do not have. This problem does not invalidate the use of these types of tools, but may call for the creation of processes that include additional phases after the extraction of entities or keywords.

Using generative AI to perform this interpretative, standardisation task (which, in theory, it is well placed to do) would also introduce significant ethical and interpretative risks into the process. Hallucination remains a risk with GenAI models, as they draw on large, uneven, and biased training data (Huang et al. 2025; Adel & Alani 2025; Alansari & Luqman 2025). The introduction of inaccurate and overgeneralised keywords into the TFH archive would be highly problematic, as it consists of user-generated stories and digitised objects contributed by people who are often still living and who shared deeply personal and sometimes painful family histories in order to honour relatives and preserve their experiences. The risks are illustrated by the original TFH workflow, where the presence of "Northern Ireland" in a record triggered the assignment of both that keyword and associated keywords within the controlled vocabulary including "Britain", "British", "Europe", "European", and "Irish", none of which existed in the contributor's text. While that workflow supported FAIR principles and metadata standards, examples such as this show how associated or inferred keywords can collapse nuanced and historically contested identities (Walker 2008), which is especially sensitive in contexts where preserving the integrity of contributor narratives is a key responsibility.

## Stewardship and the responsibility gap

Automated keyword extraction in crowdsourced collections is as much a question of responsible data stewardship as a technical or descriptive task. The CaSDaR (Careers and Skills for Data-driven Research) Network, launched in September 2025, emphasises that good data stewardship is guided by FAIR principles and supports trusted, reproducible research.[32] Equally, as Rema Patel (2018) notes, "the most compelling feature of stewardship is that it describes a good or an asset that is held on behalf of, or in trust for others", and this concept has embedded within it responsibilities and obligations towards others (Patel 2018). While in some digital collections contexts these responsibilities primarily relate to the relationship between institutions and collection users (Gooch et al. 2025), in crowdsourced collections this also concerns the contributors and families whose stories, memories, and objects form the collection. In conjunction with FAIR principles, the CARE principles (Carroll et al. 2020), which emphasise collective benefit, authority to control, responsibility, and ethics, can offer a useful perspective for those navigating the responsible use of digital technologies within crowdsourced collections contexts (Conisbee 2025).

The advent of disruptive technologies such as generative AI is increasingly challenging our ability to adjudicate moral responsibilities, resulting in a "responsibility gap" (Matthias 2004; Ayling and Chapman 2022; Dong and Bocian 2024; Fleisher et al. 2025). Building on the work of Matthias (2004) and Goetze (2021), Tollon and Vallor (2026) offer an analogy for understanding how the responsibility gap might be bridged: if a cat bites someone, its owner remains vicariously responsible for that unexpected action. Just as a pet owner is vicariously responsible for its actions, the act of choosing to deploy an AI model as part of process automation arguably creates a similar relationship of moral entanglement, with the one who chooses and deploys the model becoming answerable for resulting outcomes. Tollon and Vallor emphasise that answerability involves more than explanation: it is grounded in relationships of trust and requires the demonstration of moral concern towards those affected by any harms. As Floridi (2017) argues, there is "no ethics without choices, responsibilities, and moral evaluations, all of which need a lot of relevant information and quite good management of it". In the context of crowdsourced collections such as TFH, where responsible stewardship depends on the ability of the project team to account for how submissions data held in trust for contributors is handled and processed, there are significant implications for how AI models and approaches to support automation are selected.

Project findings indicate that (open) NER and KE models have the potential to better satisfy the conditions for responsible data stewardship than the alternatives evaluated, and these are identified as the strongest starting point for teams seeking to extract keywords at scale in similar crowdsourced collections contexts. When open NER models are used to generate keywords, those managing digital collections can be more meaningfully answerable for the models' output and more readily fulfil their data stewardship responsibilities, as resulting outputs are possible to trace, verify, and reproduce. Although the NER models tested produced omissions (entities not extracted), misclassifications (Florence the person, identified as Florence the location), and false positives (words incorrectly identified as entities), NER-generated keywords are typically traceable to the source text, as in the case of the TFH archive, as keywords are displayed alongside source text. This is in contrast to generative AI models that may produce keywords that appear plausibly correct, but the rationale for their inclusion cannot always be explained, due to the inherently probabilistic, "black box" nature of such models (Council 2023; Ayling and Chapman 2022). As a result, the person deploying AI-generated keywords is placed in a precarious position, being vicariously responsible for outputs they cannot fully answer for. Choosing a model or approach that undermines one's ability to be answerable for the results of their use can risk eroding the very trust upon which crowdsourced collections depend.

[32] https://casdar.ac.uk/about/

Importantly, using open NER or KE models to extract keywords can also better support the retention of contributors' agency, as defined by Murray (2017), by avoiding the imposition of standardised naming conventions and the normalisation of variant expressions, such as "pa", "father" and "da", reducing the homogenisation of language (and thus culture and identities) across a collection which can be ethically problematic (Bowker and Star 2000; Cocq 2022; Breemen and Breemen, 2025). The TFH project illustrated, in practice, that retention of agency over representation matters to contributors. Since the publication of the TFH archive on 6 June 2024, the project has received 66 change requests from contributors, ranging from the addition of photographs, the correction of name spellings, and requests to use nicknames in place of given names or vice versa, thus reinforcing the relevance of responsible stewardship and answerability. Still, as already stated, no model or approach is perfect, and there are certainly issues with NER and KE; being honest and open about these is just as critical for responsible stewardship. For instance, due to limited resources and scope, the project did not have the opportunity to thoroughly scrutinise the training data used by the open NER models tested, and there are important ethical considerations related to this. Further research assessing the extent to which NER training corpora encode assumptions and biases into models and how this impacts culturally diverse collections would be valuable.

At the same time, using NER and KE for keyword extraction can prioritise contributor agency over the optimisation of search and discovery for collections for users. Stewards of crowdsourced collections have a responsibility to both contributors *and* users. The FAIRness of the data and utilising platforms designed to support search, discovery, and interoperability can in some ways help fulfil this responsibility by supporting export for collections-as-data research (Padilla, 2024). More transparent processes of metadata extraction can also better support information seekers, such as historians, in applying the critical lenses they bring to traditional documents to digital collections (Fickers and Zaagsma, 2022; Ridge 2025).

## The value of interdisciplinary collaboration

Another key finding that emerged from the EKCC project is the value of collaboration with technical experts. Humanities and GLAM projects have long sought to work with technical collaborators to address problems beyond the expertise of core project teams (Gooch and Strange 2023), and the collaboration with a Research Software Engineer (RSE) was vital for this project. It enabled the team to explore NER, KE, and TM approaches more effectively, and to responsibly evaluate a wider range of model implementations. Just as importantly, it introduced beneficial friction into the process: assumptions about what different models and approaches were capable of were challenged, and the appropriateness of model choices was questioned. This is important, as while generative AI has made advanced digital methods more accessible to researchers without formal technical training , access facilitated via chat interfaces such as ChatGPT does not guarantee informed use (Viola 2023; Beavan et al. 2025; Ridge 2025). Equally, the availability of millions of public models through platforms such as Hugging Face can make model selection difficult, especially where project teams are unfamiliar with differences in training data, significance of context windows on performance, and expected outputs. [33] Collaboration with RSEs or other technical experts can support more context-appropriate, efficient, and responsible use of AI in crowdsourced collections (Viola 2023), and aid alignment with minimal computing approaches that encourage critical evaluation of purpose before technological choice (Risam and Gil 2022).

From an RSE's perspective, it is extremely valuable to be able to work closely with researchers who are experts in the field for which the technology is being developed, whilst it is being developed. From a development perspective, this provides access to expert knowledge in the field being researched, which has an impact on every development cycle. This ensures that the development remains aligned at all times with the research objectives, allows for the creation of methodologies informed by domain knowledge rather than naive approaches based solely on technical expertise, and greatly streamlines the development process as it embeds feedback from tool users directly into the development itself, rather than creating sequential phases between the two. From a research perspective, for the RSE these collaborations entail in-depth learning in the subject area, which considerably improves any future technology produced in this and any related field. It offers privileged access to this learning, which would otherwise be discouraged within the RSE environment, as it is not purely technical knowledge, and due to the high pressure to stay up to date in a technical field that is changing radically faster than experts can keep up with.

[33] https://huggingface.co/blog/huggingface/state-of-os-hf-spring-2026

Truly interdisciplinary projects, which involve joint work characterised by constant communication and co-creation among the various members of the collaboration throughout the entire project, not only produce more comprehensive and robust outputs, but also produce better researchers—who become more expert in the technical subject matter—and better technicians—who become more expert in the field under investigation. And beyond academic efficiency, these projects are far more personally enriching and enjoyable, which is no small matter in any human endeavour.

## 6. Conclusion

In this project, we investigated the use of AI techniques to address the technical, practical and ethical challenges of keyword extraction and assignment in a crowdsourced collections context. In this article, we reported the findings of the EKCC project, which used the Their Finest Hour (TFH) Archive as a case study to carry out a comprehensive exploration of automated methods across three methodological areas: KE, NER, and TM. By evaluating traditional statistical methods, neural networks, and modern generative AI models we sought to understand how automation might support keyword extraction at scale while respecting the metadata provided by living contributors.

The characteristics of crowdsourced collections such as the TFH Archive which make them an especially valuable resource, are precisely what makes them challenging to work with. First, the crowdsourced metadata is highly complex, being inconsistent in detail, length, and focus, posing a number of technical challenges for processing. Secondly, in crowdsourced collections contexts there is a direct responsibility towards contributors, and towards any outputs generated based on their contributions, such as keywords. As such these interconnected challenges had to be confronted together rather than considered in isolation, leading the project team to complete both a quantitative and qualitative evaluation of the results of testing, accompanied by an extensive discussion of related practical and ethical implications.

Our technical findings include the identification of KE and NER methodologies capable of achieving high performance metrics in the evaluated sample (96% for KE precision @ 5, and 96% and 89% for NER precision and recall). These results were achieved using statistical methods and open-weight specialised models, which offer advantages in terms of transparency, reproducibility, privacy, explainability, and low computational resource requirements. Beyond this, open zero-shot models such as GLiNER also offer the advantage of flexibility, as they address the fixed-category limitations of other NER models tested.

Our practical and ethical findings indicate that there is no single model or technology suitable for all use cases and collections. Whilst these technologies can reduce the manual labour required for some tasks, they can generate new labour requirements, including an initial assessment of the model being deployed and human review of outputs they produce, all of which significantly benefit from enhanced digital literacy and, if possible, interdisciplinary collaboration with technical experts (such as RSEs). Using AI to automate keyword extraction in crowdsourced collections arguably creates a relationship of moral entanglement and vicarious responsibility, with those using AI models being answerable for the outputs models produce. On this basis, we found NER and KE techniques with specialised local models offer a useful starting point in crowdsourced collection contexts, because of their extractive capacity for outputting text-grounded keywords that support responsible deployment and responsible stewardship. Generative AI models, due to their abstractive capabilities, may produce outputs that can enhance the discoverability of collections in different ways, but at the cost of traceability and answerability. The decision to use one set of techniques over another involves trade-offs across multiple technical, practical, and ethical considerations, and therefore requires detailed analysis and careful evaluation in relation to each collection and use case.

The EKCC project also identifies clear areas for further research. Future work should test the indicative findings on larger datasets, scrutinise the training corpora of open KE, NER, and TM models, and explore whether, and how, specific extractive keywords can be connected to the broader abstractive concepts that users might wish to search for, without compromising responsible stewardship, transparency, data integrity and responsible deployment of AI. This question may become more pressing if GLAM digital infrastructure, and digital collections users, move towards search based on natural language queries. The wider challenge, therefore, is to develop approaches to AI-supported metadata enhancement and AI-supported discovery that improve access to crowdsourced collections while preserving the relationships of trust on which such collections depend.

## Acknowledgments

We wish to acknowledge the contributors who generously shared their stories and family experiences as part of the original Their Finest Hour Project, as well as the hundreds of volunteers who played a significant role in gathering and sharing contributions as part of the Digital Collection Days.

Gratitude also to Professor Stuart Lee (PI), Andrew Cusworth and Digital Scholarship @ Oxford colleagues, as well as members of the EKCC project Steering Group.

## Funding

The Their Finest Hour project (2022-2024) was generously supported with funding from the National Lottery Heritage Fund.

Extracting Keywords from Crowdsourced Collections project (2024-25) was generously supported with funding Digital Scholarship @ Oxford, University of Oxford.

## References

Abdelrazek, A., Eid, Y., Gawish, E., Medhat, W., & Hassan, A. (2023). Topic modeling algorithms and applications: A survey. Information Systems, 112, 102131.

Adel, A., & Alani, N. (2025). Can generative AI reliably synthesise literature? exploring hallucination issues in ChatGPT. AI & SOCIETY, 1-14.

Aejas, B., Bouras, A., Belhi, A., Gasmi, H. (2021). Named Entity Recognition for Cultural Heritage Preservation. In: Belhi, A., Bouras, A., Al-Ali, A.K., Sadka, A.H. (eds) Data Analytics for Cultural Heritage. Springer, Cham. https://doi.org/10.1007/978-3-030-66777-1_11

Akbik, A., Blythe, D., & Vollgraf, R. (2018, August). Contextual string embeddings for sequence labeling. In Proceedings of the 27th international conference on computational linguistics (pp. 1638-1649).

Alami Merrouni, Z., Frikh, B., & Ouhbi, B. (2020). Automatic keyphrase extraction: a survey and trends. Journal of Intelligent Information Systems, 54(2), 391-424.

Alansari, A., & Luqman, H. (2025). Large language models hallucination: A comprehensive survey. arXiv preprint arXiv:2510.06265.

Andresel, M., Gordea, S., Stevanetic, S., & Schütz, M. (2023). An approach for curating collections of historical documents with the use of topic detection technologies. International Journal of Digital Curation, 17(1), 12-12.

Arana-Catania, M., Conisbee, C., Kidd, M., & Lee, S. (2025). Crowdsourced data tools. University of Oxford. https://doi.org/10.5287/ora-yqqxywn02

Arthur, P. (2014). Memory and commemoration in the digital present. In Contemporary approaches in literary trauma theory (pp. 152-175). London: Palgrave Macmillan UK. https://doi.org/10.1057/9781137365941

Ayling, J., & Chapman, A. (2022). Putting AI ethics to work: are the tools fit for purpose?. AI and Ethics, 2(3), 405-429. https://doi.org/10.1007/s43681-021-00084-x

Bailey, R. A., Pereda, J., Michaels, C., & Callahan, T. (2024). Unlocking the Potential of Digital Collections. A call to action. Arts and Humanities Research Council. https://doi.org/10.5281/zenodo.13838916

Beavan, D., Piza, A., Rob, H., Stephanie, F., Francois, P., & Emma, R. (2025). Evidencing the Impact of Research Software Engineers on Arts & Humanities Scholarship. Zenodo. https://doi.org/10.5281/zenodo.15083632

Bell, S. (2025) Welcome to the new National Archives online catalogue https://www.nationalarchives.gov.uk/blogs/digital/new-online-catalogue-blog/welcome-to-the-new-national-archives-online-catalogue/ Accessed May 2026

Belteki, D., Rees, A.J. and Sichani, A.-M. (2025) 'Datafication and Cultural Heritage Collections Data Infrastructures: Critical Perspectives on Documentation, Cataloguing and Data-sharing in Cultural Heritage Institutions', Journal of Open Humanities Data, 11(1), p. 14.

Berglund Prytz, Y. (2021). Crowdsourcing and Community Collections. Retrieved from source https://blogs.it.ox.ac.uk/acit-rs-team/about/gathering-research-data/crowdsourcing/ (last accessed: 1 February 2026).

Blei, D. M., Ng, A. Y., & Jordan, M. I. (2003). Latent dirichlet allocation. Journal of machine Learning research, 3(Jan), 993-1022.

Branley-Bell, D., Feiner, J., Prossnegg, S., & Libal, T. (2025). AI and the Human Being–The Human in the Loop and AI Literacy within the AI Act. Jusletter IT, 2025, 83-92.

Boon, T., Butterworth, A., Graham, H., Rees, A., Sichani, A.-M., Belteki, D., Fitzpatrick, A., & Winters, J. (2025). Final Report - Congruence Engine: Digital Tools for New Collections-Based Industrial Histories. Towards a National Collection. https://doi.org/10.5281/zenodo.14770779

Boudin, F. (2016, December). PKE: an open source python-based keyphrase extraction toolkit. In Proceedings of COLING 2016, the 26th international conference on computational linguistics: system demonstrations (pp. 69-73).

Boudin, F. (2018, June). Unsupervised keyphrase extraction with multipartite graphs. In Proceedings of the 2018 conference of the North American chapter of the Association for Computational Linguistics: Human language technologies, volume 2 (short papers) (pp. 667-672).

Bougouin, A., Boudin, F., & Daille, B. (2013, October). Topicrank: Graph-based topic ranking for keyphrase extraction. In Proceedings of the sixth international joint conference on natural language processing (pp. 543-551).

Bowker, G. C., & Star, S. L. (2000). Sorting things out: Classification and its consequences. MIT press.

Breemen, K., & Breemen, V. (2025). 'Slow libraries' and 'Cultural AI': Reassessing technology regulation in the context of digitised cultural heritage data. Technology and Regulation, 2025, 175-193.

Lu, C., Park, J. R., & Hu, X. (2010). User tags versus expert-assigned subject terms: A comparison of LibraryThing tags and Library of Congress Subject Headings. Journal of information science, 36(6), 763-779.

Cain, J. O. (2016). Using Topic Modeling to Enhance Access to Library Digital Collections. Journal of Web Librarianship, 10(3), 210–225. https://doi.org/10.1080/19322909.2016.1193455

Campos, R., Mangaravite, V., Pasquali, A., Jorge, A. M., Nunes, C., & Jatowt, A. (2018a, March). A text feature based automatic keyword extraction method for single documents. In European conference on information retrieval (pp. 684-691). Cham: Springer International Publishing.

Campos, R., Mangaravite, V., Pasquali, A., Jorge, A. M., Nunes, C., & Jatowt, A. (2018b, March). Yake! collection-independent automatic keyword extractor. In European conference on information retrieval (pp. 806-810). Cham: Springer International Publishing.

Campos, R., Mangaravite, V., Pasquali, A., Jorge, A., Nunes, C., & Jatowt, A. (2020). YAKE! Keyword extraction from single documents using multiple local features. Information Sciences, 509, 257-289.

Carroll, S. R., Garba, I., Figueroa-Rodríguez, O. L., Holbrook, J., Lovett, R., Materechera, S., Parsons, M., Raseroka, K., Rodriguez-Lonebear, D., Rowe, R., Rodrigo, S., Walker, J., Anderson, J., & Hudson, M. (2020).

The CARE Principles for Indigenous Data Governance. Data Science Journal, 19(1), 43. https://doi.org/10.5334/dsj-2020-043

Carter, K. S., Gondek, A., Underwood, W., Randby, T., & Marciano, R. (2022). Using AI and ML to optimize information discovery in under-utilized, Holocaust-related records. AI & SOCIETY, 37(3), 837-858. https://doi.org/10.1007/s00146-021-01368-w

CaSDaR (n.d.) Why Data Stewardship https://casdar.ac.uk/why-data-stewardship/ Accessed May 2026

Cebr. (2025) The economic impact of closing the work essential digital skills gap. https://futuredotnow.uk/wp-content/uploads/2025/05/Final-Report-The-economic-impact-of-closing-the-work-essential-digital-skills-gap.pdf Accessed May 2026

Cevolini, A. (2026). Indexing Systems: The Evolution of Knowledge Organisation in Modern Society (pp. 1-220). Palgrave Macmillan.

Chou, C., & Chu, T. (2022). An Analysis of BERT (NLP) for Assisted Subject Indexing for Project Gutenberg. Cataloging & Classification Quarterly, 60(8), 807–835. https://doi.org/10.1080/01639374.2022.2138666

Churchill, R., & Singh, L. (2022). The evolution of topic modeling. ACM Computing Surveys, 54(10s), 1-35. https://doi.org/10.1145/3507900

Cocq, C. (2022). Revisiting the digital humanities through the lens of Indigenous studies—or how to question the cultural blindness of our technologies and practices. Journal of the Association for Information Science and Technology, 73(2), 333-344. https://doi.org/10.1002/asi.24564

Colla, D., Goy, A., Leontino, M., Magro, D., & Picardi, C. (2022). Bringing semantics into historical archives with computer-aided rich metadata generation. Journal on Computing and Cultural Heritage (JOCCH), 15(3), 1-24. https://doi.org/10.1145/3484398

Colavizza, G., Blanke, T., Jeurgens, C., & Noordegraaf, J. (2021). Archives and AI: An overview of current debates and future perspectives. ACM Journal on Computing and Cultural Heritage (JOCCH), 15(1), 1-15.

Conisbee, C. (2025). Research Data Management and Crowdsourcing Personal Histories. Journal of Open Humanities Data, 11(1). https://doi.org/10.5334/johd.265

Council, N. C. (2023). Ghost in the Machine-Addressing the Consumer Harms of Generative AI. Norwegian Consumer Council, June. https://storage02.forbrukerradet.no/media/2023/06/generative-ai-rapport-2023.pdf

Dong, M., & Bocian, K. (2024). Responsibility gaps and self-interest bias: People attribute moral responsibility to AI for their own but not others' transgressions. Journal of Experimental Social Psychology, 111, 104584. https://doi.org/10.1016/j.jesp.2023.104584

Drabczyk, M., Rendina, M., Manfredini, F., Dimou, A., Peeters, R., de Koning, J., & Svorc, J. (2025). Comprehensive Guide of the Benefits, Opportunities, Risks and Gaps in the Management of Cultural Heritage Digitisation. A Critical Literature Review. 3D Research Challenges in Cultural Heritage VI: DigitalTwin versus MemoryTwin, 102-111. https://doi.org/10.1007/978-3-032-05656-6_10

Ehrmann, M., Hamdi, A., Pontes, E. L., Romanello, M., & Doucet, A. (2023). Named entity recognition and classification in historical documents: A survey. ACM Computing Surveys, 56(2), 1-47. https://doi.org/10.1145/3604931

El-Beltagy, S. R., & Rafea, A. (2010, July). Kp-miner: Participation in semeval-2. In Proceedings of the 5th international workshop on semantic evaluation (pp. 190-193).

Eshima, S., Imai, K. and Sasaki, T. (2024), Keyword-Assisted Topic Models. American Journal of Political Science, 68: 730-750. https://doi.org/10.1111/ajps.12779

Fickers, A., & Zaagsma, G. (2022). Digital hermeneutics: The reflexive turn in digital public history. Handbook of Digital Public History. Berlin: De Gruyter, 139-48.

Firoozeh, N., Nazarenko, A., Alizon, F., & Daille, B. (2020). Keyword extraction: Issues and methods. Natural Language Engineering, 26(3), 259-291.

Fleisher, W., Cibralic, B., Basl, J., Ricks, V., & Smith, M. N. (2025). Responsibility and accountability in an algorithmic society. Philosophy & Technology, 38(4), 144. https://doi.org/10.1007/s13347-025-00970-w

Florescu, C., & Caragea, C. (2017, July). Positionrank: An unsupervised approach to keyphrase extraction from scholarly documents. In Proceedings of the 55th annual meeting of the association for computational linguistics (volume 1: long papers) (pp. 1105-1115).

Floridi, L. (2017) https://www.thenewatlantis.com/publications/why-information-matters Accessed May 2026

Giarelis, N., & Karacapilidis, N. (2024). Deep learning and embeddings-based approaches for keyphrase extraction: a literature review. Knowledge and Information Systems, 66(11), 6493-6526.

Glowacka-Musial, M. (2022). Applying Topic Modeling for Automated Creation of Descriptive Metadata for Digital Collections. Information Technology and Libraries, 41(2). https://doi.org/10.6017/ital.v41i2.13799

Goetze, T. S. (2021). Moral entanglement: Taking responsibility and vicarious responsibility. The Monist, 104(2), 210-223.

Golub, K. (2021). Automated Subject Indexing: An Overview. Cataloging & Classification Quarterly, 59(8), 702–719. https://doi.org/10.1080/01639374.2021.2012311

Golub, K., Soergel, D., Buchanan, G., Tudhope, D., Lykke, M. and Hiom, D. (2016), A framework for evaluating automatic indexing or classification in the context of retrieval. J Assn Inf Sci Tec, 67: 3-16. https://doi.org/10.1002/asi.23600

Gooch, M., & Strange, D. (2023). Consolidating research data management infrastructure: Towards sustainable digital scholarship. ACM Journal on Computing and Cultural Heritage, 16(4), 1-16.

Gooch, M., Kahn, R., & Kugeler, E. (2025). Human perspectives and social infrastructures: prioritising people in GLAM digitisation. Journal of Open Humanities Data, 11.

Gosling, K., McKenna, G., & Cooper, A. (2022). Digital collections audit. Zenodo. https://doi.org/10.5281/zenodo.6379581

Grootendorst, M (2021). MaartenGr/KeyBERT: BibTeX (v0.1.3). Zenodo. https://doi.org/10.5281/zenodo.4461265

Grootendorst, M. (2022). BERTopic: Neural topic modeling with a class-based TF-IDF procedure. arXiv preprint arXiv:2203.05794.

Gross, T., Taylor, A. G., & Joudrey, D. N. (2015). Still a Lot to Lose: The Role of Controlled Vocabulary in Keyword Searching. Cataloging & Classification Quarterly, 53(1), 1–39. https://doi.org/10.1080/01639374.2014.917447

Haffenden, C., Fano, E., Malmsten, M., & Börjeson, L. (2023). Making and using AI in the library: Creating a BERT model at the National Library of Sweden. College & research libraries, 84(1).

Hagedorn, K., Chapman, S., Newman, D., & Irvine, C. A. (2007). Enhancing search and browse using automated clustering of subject metadata. D-Lib Magazine, 13(7/8), 1082-9873.

Harandizadeh, B., Priniski, J. H., & Morstatter, F. (2022, February). Keyword assisted embedded topic model. In Proceedings of the fifteenth ACM international conference on web search and data mining (pp. 372-380).

Hernandez-Camero, I., Garcia-Cabot, A., Garcia-Lopez, E., Caro-Alvaro, S., & Moreno-Cediel, A. (2025). Leveraging large language models to enhance clustering-based topic modeling. Knowledge and Information Systems, 67(12), 12661-12697. https://doi.org/10.1007/s10115-025-02605-0

Hoffman, M., Bach, F., & Blei, D. (2010). Online learning for latent dirichlet allocation. advances in neural information processing systems, 23.

Honnibal, M., Montani, I., Van Landeghem, S., & Boyd, A. (2020). spaCy: Industrial-strength natural language processing in Python.

Hoyer, P. O. (2004). Non-negative matrix factorization with sparseness constraints. Journal of machine learning research, 5(Nov), 1457-1469.

Huang, L., Yu, W., Ma, W., Zhong, W., Feng, Z., Wang, H., ... & Liu, T. (2025). A survey on hallucination in large language models: Principles, taxonomy, challenges, and open questions. ACM Transactions on Information Systems, 43(2), 1-55.

Hughes, L., Alexander, M., Barker, H., Batista-Navarro, R., Nenadic, G., Sheridan, J., Hannaford, E., & Dierckx, K. (2025). Final Report - Our Heritage, Our Stories: Linking and searching Community-Generated Digital Content. Towards a National Collection. https://doi.org/10.5281/zenodo.14888428

Jaillant, Lise. "Introduction." Archives, Access and Artificial Intelligence, edited by Lise Jaillant, vol. 2, Bielefeld University Press, 2022, pp. 7–28, https://doi.org/10.1515/9783839455845-001.

Keraghel, I., Morbieu, S., & Nadif, M. (2024). Recent advances in named entity recognition: A comprehensive survey and comparative study. arXiv preprint arXiv:2401.10825.

Kipp, M. E., & Campbell, D. G. (2010). Searching with tags: do tags help users find things?. Knowledge organization, 37(4). https://doi.org/10.5771/0943-7444-2010-4-239

Korn, N., Waterman, S., & Associates, NK. (2024) Changes and Challenges in Heritage and Open Knowledge. https://www.culturehive.co.uk/digital-heritage-hub/resource/leadership/heritage-open-knowledge/ Accessed May 2026

Lazaros, K., Vrahatis, A. G., & Kotsiantis, S. (2026). Human-in-the-Loop Artificial Intelligence: A Systematic Review of Concepts, Methods, and Applications. Entropy, 28(4), 377.

Lee, D., & Seung, H. S. (2000). Algorithms for non-negative matrix factorization. Advances in neural information processing systems, 13.

Lee, S., Kidd, M., Carrington, N., Conisbee, C., & Quinn, J. (2024). Their Finest Hour Online Archive [Dataset]. University of Oxford. https://doi.org/10.5287/ora-nzpbbm0v5

Long, K., Thompson, S., Potvin, S., & Rivero, M. (2017). The "wicked problem" of neutral description: Toward a documentation approach to metadata standards. Cataloging & Classification Quarterly, 55(3), 107-128. https://doi.org/10.1080/01639374.2016.1278419

Long, S., & Li, W. (2026). A Chinese Named Entity Recognition Dataset for Intangible Cultural Heritage. Scientific Data. https://doi.org/10.1038/s41597-026-06700-x

Luthra, M., Todorov, K., Jeurgens, C., & Colavizza, G. (2024). Unsilencing colonial archives via automated entity recognition. Journal of Documentation, 80(5), 1080-1105. https://doi.org/10.1108/JD-02-2022-0038

Macgregor, G., & McCulloch, E. (2006). Collaborative tagging as a knowledge organisation and resource discovery tool. Library review, 55(5), 291-300.

Maragheh, R. Y., Fang, C., Irugu, C. C., Parikh, P., Cho, J., Xu, J., ... & Achan, K. (2023, December). LLM-TAKE: Theme-aware keyword extraction using large language models. In 2023 IEEE International Conference on Big Data (BigData) (pp. 4318-4324). IEEE. doi: 10.1109/BigData59044.2023.10386476.

Massey, J., Moriniere, S., Parkes, E., Adigun, Y., Crowe, J. & Hardinges, J. (2023) Defining responsible data stewardship https://theodi.org/insights/reports/defining-responsible-data-stewardship/ Accessed May 2026

Matthias, A. (2004). The responsibility gap: Ascribing responsibility for the actions of learning automata. Ethics and information technology, 6(3), 175-183. https://doi.org/10.1007/s10676-004-3422-1

Mergel, I., Kühler, J., Schumann, AL., Nitsch, C. (2026). Automated Subject Cataloguing at the German National Library. In: Mergel, I., Schmidt, C. (eds) AI Innovations in Public Services. Springer, Cham. https://doi.org/10.1007/978-3-032-01344-6_3

Alami Merrouni, Z., Frikh, B., & Ouhbi, B. (2020). Automatic keyphrase extraction: a survey and trends. Journal of Intelligent Information Systems, 54(2), 391-424.

Mihalcea, R., & Tarau, P. (2004, July). Textrank: Bringing order into text. In Proceedings of the 2004 conference on empirical methods in natural language processing (pp. 404-411).

Murray, J. H. (2017). Hamlet on the Holodeck, updated edition: The Future of Narrative in Cyberspace. MIT press.
http://itgs.pbworks.com/w/file/fetch/143077635/Murray%20-%20Hamlet%20on%20the%20Holodeck%20The%20Future%20of%20Narrative%20in%20Cyberspace.pdf

Newman, D., Noh, Y., Talley, E., Karimi, S., & Baldwin, T. (2010, June). Evaluating topic models for digital libraries. In Proceedings of the 10th annual joint conference on Digital libraries (pp. 215-224) https://doi.org/10.1145/1816123.1816156

Newman-Griffis, D. (2025). AI thinking: a framework for rethinking artificial intelligence in practice. Royal Society Open Science, 12(1), 241482. https://doi.org/10.1098/rsos.241482

Paatero, P., & Tapper, U. (1994). Positive matrix factorization: A non-negative factor model with optimal utilization of error estimates of data values. Environmetrics, 5(2), 111-126.

Pacheco, A., Da Silva, C. G., & De Freitas, M. C. V. (2023). A metadata model for authenticity in digital archival descriptions. Archival Science, 23(4), 629-673. https://doi.org/10.1007/s10502-023-09422-w

Padilla, T., Scates Kettler, H., & Shorish, Y. (2023). Collections as Data: Part to Whole Final Report (Version 1). Zenodo. https://doi.org/10.5281/zenodo.10161976

Padmavilochanan, A., Gangadharan, V., Rashed, T., & Natarajan, A. (2025). Analyzing Vulnerability Through Narratives: A Prompt-Based NLP Framework for Information Extraction and Insight Generation. Big Data and Cognitive Computing, 10(1), 6. https://doi.org/10.3390/bdcc10010006

Patel, R. (2018). Data for the common good – framing the debate. https://www.adalovelaceinstitute.org/blog/data-for-the-common-good-framing-the-debate/ Accessed 10 March 2026

Pedregosa, F., Varoquaux, G., Gramfort, A., Michel, V., Thirion, B., Grisel, O., ... & Duchesnay, É. (2011). Scikit-learn: Machine learning in Python. the Journal of machine Learning research, 12, 2825-2830.

Pirniakan, M., Adibifar, A., Litooee, F. S., Tangestanizadeh, M., & Etebar Zadeh, M. (2026). Enhancing topic modeling in digital libraries through keyword-centric self-supervised learning. Journal of Librarianship and Information Science, 58(2), 733-749.

Poley, C., Uhlmann, S., Busse, F., Jacobs, J.-H., Kähler, M., Nagelschmidt, M., & Schumacher, M. (2025). Automatic Subject Cataloguing at the German National Library. LIBER Quarterly: The Journal of the Association of European Research Libraries, 35(1), 1-29. https://doi.org/10.53377/lq.19422

Poole, A. H. (2019). Social tagging and commenting in participatory archives: a critical literature review. Participatory archives: theory and practice, 15-31.

Ramsden, S. (2023). Community Generated Digital Content and the Postmemory of the Second World War - Our Heritage, Our Stories blog series. Zenodo. https://doi.org/10.5281/zenodo.11244081

Reimers, N., & Gurevych, I. (2019, November). Sentence-bert: Sentence embeddings using siamese bert-networks. In Proceedings of the 2019 conference on empirical methods in natural language processing and the 9th international joint conference on natural language processing (EMNLP-IJCNLP) (pp. 3982-3992).

Reusens, M., Adams, A., & Baesens, B. (2025). Large Language Models to make museum archive collections more accessible. AI & SOCIETY, 40(6), 4485-4497. https://doi.org/10.1007/s00146-025-02227-8

Ridge, M. (2025). Seeing Library Collections Through New Lenses: The Potential of Large-Scale Digital Collections. Journal of Victorian Culture, vcaf021. https://doi.org/10.1093/jvcult/vcaf021

Ridge, M., Blickhan, S., Ferriter, M., Mast, A., Brumfield, B., Wilkins, B., ... Prytz, Y. B. (2021). 8. Choosing tasks and workflows. In The Collective Wisdom Handbook: Perspectives on Crowdsourcing in Cultural Heritage - community review version (1st ed.). https://doi.org/10.21428/a5d7554f.1b80974b

Ridge, M., Ferriter, M., & Blickhan, S. (2023). Recommendations, challenges and opportunities for the future of crowdsourcing in cultural heritage: a White Paper. Digital Scholarship at the British Library. https://doi.org/10.21428/a5d7554f.2a84f94b

Risam, R., & Gil, A. (2022). Introduction: The Questions of Minimal Computing. DHQ: Digital Humanities Quarterly, 16(2).

Ranade, S. (2016, December). Traces through time: A probabilistic approach to connected archival data. In 2016 IEEE International Conference on Big Data (Big Data) (pp. 3260-3265). IEEE. doi: 10.1109/BigData.2016.7840983.

Salse, M., Guallar-Delgado, J., Jornet-Benito, N., Mateo Bretos, M. P., & Silvestre-Canut, J. O. (2024). GLAM metadata in museums and university collections: a state-of-the-art (Spain and other European countries). Global Knowledge, Memory and Communication, 73(4-5), 477-495. https://doi.org/10.1108/GKMC-06-2022-0133

Schultz, M. D., Conti, L. G., & Seele, P. (2025). Digital ethicswashing: A systematic review and a process-perception-outcome framework. AI and Ethics, 5(2), 805-818. https://doi.org/10.1007/s43681-024-00430-9

Schumann, AL., Mergel, I. (2026). AI-Supported Automated Subject Indexing and Metadata Management with Annif and Finto AI at the National Library of Finland. In: Mergel, I., Schmidt, C. (eds) AI Innovations in Public Services. Springer, Cham. https://doi.org/10.1007/978-3-032-01344-6_4

Schwartz, J. M., & Cook, T. (2002). Archives, records, and power: The making of modern memory. Archival science, 2(1), 1-19. https://doi.org/10.1007/BF02435628

Shah, C., & Bender, E. M. (2022, March). Situating search. In Proceedings of the 2022 Conference on Human Information Interaction and Retrieval (pp. 221-232).

Škrlj, B., Koloski, B., & Pollak, S. (2022, October). Retrieval-efficiency trade-off of unsupervised keyword extraction. In International Conference on Discovery Science (pp. 379-393). Cham: Springer Nature Switzerland.

Škrlj, B., Repar, A., & Pollak, S. (2019, September). RaKUn: Ra nk-based K eyword extraction via Un supervised learning and meta vertex aggregation. In International Conference on Statistical Language and Speech Processing (pp. 311-323). Cham: Springer International Publishing.

Taka, E., Nakao, Y., Sonoda, R., Yokota, T., Luo, L., & Stumpf, S. (2023). Human-in-the-loop fairness: Integrating stakeholder feedback to incorporate fairness perspectives in responsible AI. arXiv preprint arXiv:2312.08064.

Their Finest Hour. Explore by Collection Day. (n.d.) https://theirfinesthour.english.ox.ac.uk/explore-by-collection-day Accessed May 2026

Tollon, F., & Vallor, S. (2026). Entangling ourselves with AI: Affirmative responsibility and the cultivation of responsible agency. Contemporary debates in the ethics of artificial intelligence, 161-181. https://doi.org/10.1002/9781394258840.ch12

Van Hooland, S., De Wilde, M., Verborgh, R., Steiner, T., & Van de Walle, R. (2015). Exploring entity recognition and disambiguation for cultural heritage collections. Digital Scholarship in the Humanities, 30(2), 262-279. https://doi.org/10.1093/llc/fqt067

Viola, L. (2023). The Humanities in the Digital: Beyond Critical Digital Humanities . Cham: Springer International Publishing. https://doi.org/10.1007/978-3-031-16950-2

Walker, B. (2008) British or Irish - who do you think you are? https://www.belfasttelegraph.co.uk/comment/british-or-irish-who-do-you-think-you-are/a/119226374.html Accessed May 2026

Wan, X., & Xiao, J. (2008, August). CollabRank: towards a collaborative approach to single-document keyphrase extraction. In Proceedings of the 22nd International Conference on Computational Linguistics (Coling 2008) (pp. 969-976).

Wang, R., Wang, Y., Li, Z., Cheng, H., & Sun, G. (2025). Distributed Keyword-guided Topic Model with Lexical Knowledge Supervision. ACM Transactions on Knowledge Discovery from Data, 19(6), 1-19. https://doi.org/10.1145/3737881

Wilkinson, M. D., Dumontier, M., Aalbersberg, I. J., Appleton, G., Axton, M., Baak, A., ... & Mons, B. (2016). The FAIR Guiding Principles for scientific data management and stewardship. Scientific data, 3(1), 1-9. https://doi.org/10.1038/sdata.2016.18

Winner, L. (1986). The whale and the reactor: A search for limits in an age of high technology. University of Chicago Press.

Zaratiana, U., Tomeh, N., Holat, P., & Charnois, T. (2024, June). GLiNER: Generalist model for named entity recognition using bidirectional transformer. In Proceedings of the 2024 Conference of the North American Chapter of the Association for Computational Linguistics: Human Language Technologies (Volume 1: Long Papers) (pp. 5364-5376).